%% file: main.tex
\documentclass[10pt,twocolumn,letterpaper]{article}

\usepackage{cvpr} 
\usepackage[listings]{tcolorbox}

\input{config/config}

\input{config/commands}

\input{config/params}
\usepackage[accsupp]{axessibility} 

\usepackage{xcolor, soul}
\usepackage[pagebackref=true,breaklinks=true,colorlinks,bookmarks=false]{hyperref}
\hypersetup{linkcolor=[rgb]{0.7,0.1,0.1}}
\hypersetup{citecolor=[rgb]{0.4,0.15,0.95}}
\definecolor{cvprblue}{rgb}{0.21,0.49,0.74}

\sethlcolor{yellow}

\newcommand{\model}{ChatPose\xspace}
\begin{document}
\title{ChatPose: Chatting about 3D Human Pose}
\author{\fontsize{11.5pt}{\baselineskip}\selectfont Yao Feng\textsuperscript{1,2,3} ~~Jing Lin\textsuperscript{3,4} ~~Sai Kumar Dwivedi\textsuperscript{1}
~~Yu Sun\textsuperscript{3} ~~Priyanka Patel\textsuperscript{1} ~~Michael J. Black\textsuperscript{1}\\[0.25mm]
\fontsize{11pt}{\baselineskip}\selectfont\textsuperscript{1}Max Planck Institute for Intelligent Systems - Tübingen ~~\textsuperscript{2}ETH Zürich
\\\fontsize{11pt}{\baselineskip}\selectfont\textsuperscript{3}Meshcapade~~~~\textsuperscript{4}Tsinghua University
}
\twocolumn[{%
\renewcommand\twocolumn[1][]{#1}%
\maketitle
 \vspace*{-1cm}
\begin{center}
    \centering
    \captionsetup{type=figure}
    \includegraphics[width=0.99\linewidth]{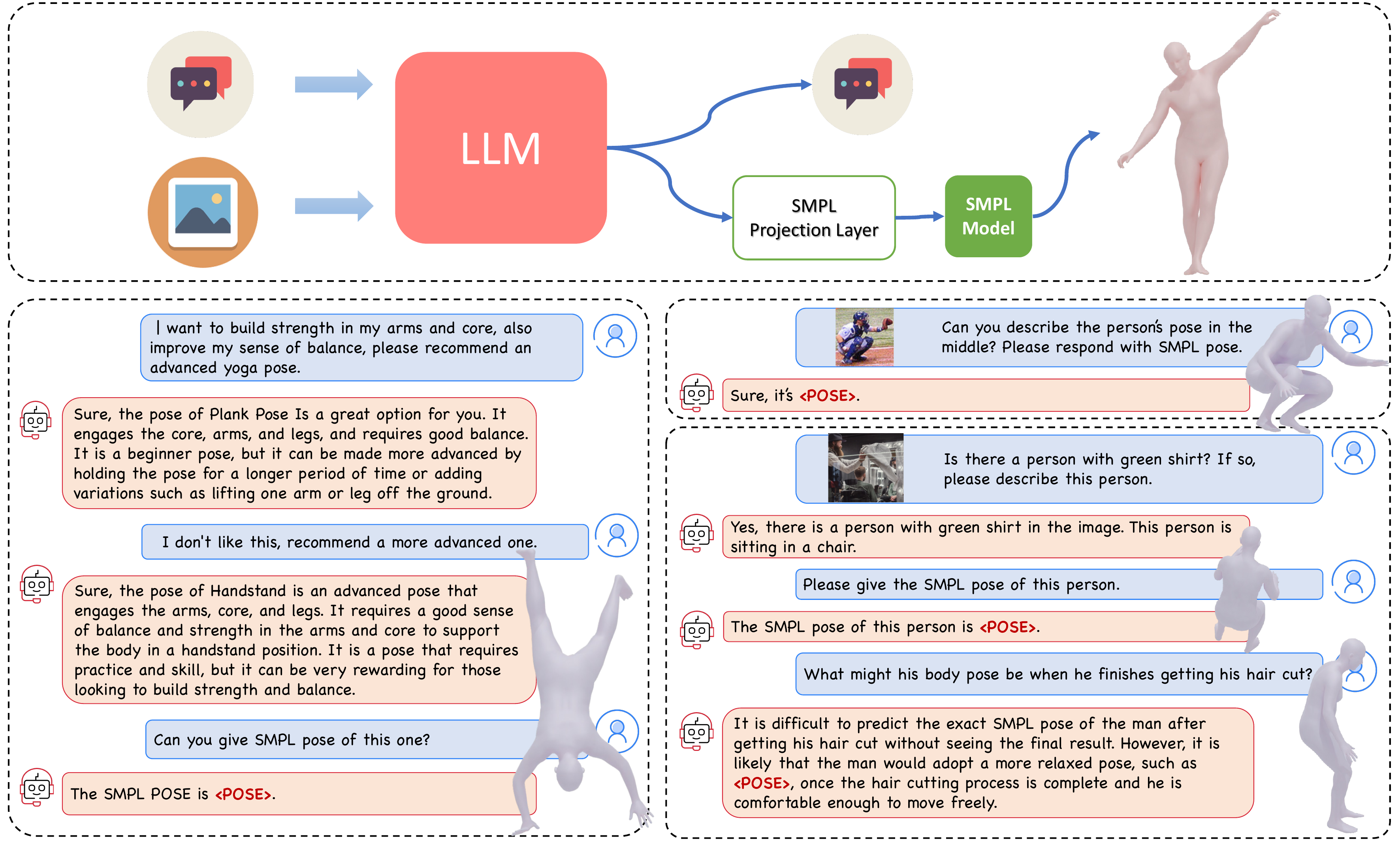}
    \captionof{figure}{We introduce \model, a multimodel LLM designed for chatting about human pose that produces 3D human poses (SMPL pose parameters) upon user request. \model features a specialized SMPL projection layer trained to convert language embeddings into 3D human pose parameters. Our demonstration includes conversations both without (left) and with (right) an image input.  Upon detection of a pose token, the token is used to estimate the SMPL pose parameters and subsequently generate the corresponding 3D body mesh. } \label{fig:teaser} 
\end{center}%
}]

\input{sections/0_abstract}    
\input{sections/1_intro}
\input{sections/2_related_work}

\input{sections/3_method}
\input{sections/4_experiment}

\input{sections/5_conclusion}
\input{sections/6_acknowledgement}
\input{sections/X_suppl} 

{
    \small
    \bibliographystyle{ieeenat_fullname}
    \bibliography{main}
}

\end{document}

%% file: config/config.tex
\usepackage{times}
\usepackage{epsfig}
\usepackage{graphicx}
\usepackage{amsmath}
\usepackage{amssymb}
\usepackage{caption}
\usepackage{enumitem}
\usepackage{booktabs}

\usepackage{caption}
\usepackage{makecell}
\usepackage{multirow}
\usepackage{xcolor, colortbl}
\usepackage{cuted}
\usepackage{capt-of}
\usepackage{comment}
\usepackage[normalem]{ulem} %
\usepackage{pifont}
\usepackage{xspace}

\usepackage{siunitx}  
\usepackage{bigstrut} 
\usepackage{array}   

\usepackage[pagebackref=true,breaklinks=true,colorlinks,bookmarks=false,pagebackref]{hyperref}

\usepackage[normalem]{ulem}

\usepackage{afterpage}

%% file: config/commands.tex
\newcommand{\methodname}{ChatPose\xspace}

\newcommand{\captionArrows}{Arrows show whether higher or lower values are better}
\newcommand{\captionBold}{Bold shows the best model for each metric}

\newcommand{\cheading}[1]{\noindent\textbf{#1.}}

\usepackage{lipsum}
\usepackage{balance}

\definecolor{citecolor}{HTML}{0071bc}
\definecolor{frontcolor}{HTML}{325ea5}
\definecolor{sidecolor}{HTML}{a58b77}
\definecolor{DeltaColor}{rgb}{0.039,0.73,0.71}
\definecolor{SigmaColor}{rgb}{0.98,0.45,0.0}
\definecolor{AlphaColor}{rgb}{0,0,0.8}
\definecolor{BetaColor}{rgb}{0.8,0,0.8}
\definecolor{GammaColor}{rgb}{0.514,0.34,0.224}
\definecolor{EpsilonColor}{rgb}{0.353,0.725,0.906}
\definecolor{PurpleColor}{HTML}{bca5ea}
\definecolor{OrangeColor}{rgb}{0.914,0.541,0.0.141}
\definecolor{GreenColor}{rgb}{0.137,0.573,0.565}
\definecolor{RedColor}{rgb}{0.949,0.275, 0.224}
\definecolor{LightCyan}{rgb}{0.88,1,1}
\definecolor{Gray}{gray}{0.85}

\newcommand{\dataColor}{black}

\newcommand{\threedpw}{\mbox{\textcolor{\dataColor}{3DPW}}\xspace}

\newcommand{\humanthreesix}{\mbox{H3.6M}\xspace}

\newcolumntype{a}{>{\columncolor{Gray}}c}

\newcommand{\CR}[1]{{\color{black} #1}}

\newcommand{\moveToSupMat}[1]{\begin{comment}#1\end{commment}}
\newcommand{\supmat}{\textcolor{black}{\emph{Sup.~Mat.}}\xspace}

\newcommand{\smplify}{\mbox{SMPLify}\xspace}

\newcommand{\llava}{\mbox{LLaVA}\xspace}
\newcommand{\posescript}{\mbox{PoseScript}\xspace}

\newcommand{\colorRef}[1]{\textcolor{red}{#1}} %
\usepackage[capitalize]{cleveref}
\crefname{figure}{\colorRef{Fig.}}{\colorRef{Figs.}}
\Crefname{figure}{\colorRef{Figure}}{\colorRef{Figures}}
\crefname{section}{\colorRef{Sec.}}{\colorRef{Secs.}}
\Crefname{section}{\colorRef{Section}}{\colorRef{Sections}}
\Crefname{table}{\colorRef{Table}}{\colorRef{Tables}}
\crefname{table}{\colorRef{Tab.}}{\colorRef{Tabs.}}
\Crefname{equation}{\colorRef{Equation}}{\colorRef{Equations}}
\crefname{equation}{\colorRef{Eq.}}{\colorRef{Eqs.}}

%% file: sections/0_abstract.tex
\begin{abstract}
We introduce ChatPose, a framework employing Large Language Models (LLMs) to understand and reason about 3D human poses from images or textual descriptions. 
Our work is motivated by the human ability to intuitively understand postures from a single image or a brief description, a process that intertwines image interpretation, world knowledge, and an understanding of body language. 
Traditional human pose estimation and generation methods often operate in isolation, lacking semantic understanding and reasoning abilities. ChatPose addresses these limitations by embedding SMPL poses as distinct signal tokens within a multimodal LLM, enabling the direct generation of 3D body poses from both textual and visual inputs. 
Leveraging the powerful capabilities of multimodal LLMs, ChatPose unifies classical 3D human pose and generation tasks while offering user interactions. 
Additionally, ChatPose empowers LLMs to apply their extensive world knowledge in reasoning about human poses, leading to two advanced tasks: speculative pose generation and reasoning about pose estimation. 
These tasks involve reasoning about humans to generate 3D poses from subtle text queries, possibly accompanied by images. We establish benchmarks for these tasks, moving beyond traditional 3D pose generation and estimation methods. 
Our results show that ChatPose outperforms existing multimodal LLMs and task-specific methods on these newly proposed tasks. Furthermore, ChatPose's ability to understand and generate 3D human poses based on complex reasoning opens new directions in human pose analysis. Code and data are available for research at \url{https://yfeng95.github.io/ChatPose}.

\end{abstract} 

%% file: sections/1_intro.tex
\section{Introduction}
\label{sec:intro}
\CR{We address the problem of understanding and reasoning about 3D human pose from an image or a text description via large language models. 
} 
For humans, a quick glance at a picture or a brief description of a person allows us to form an impression of their articulated body posture.
For instance, one might wonder, ``What is the girl in the dress doing?" or ``How might she behave if she feels tired?".  This involves interpreting the image, employing general knowledge about the world, and understanding human body language.
Current methods that estimate 3D human poses from images \cite{hmr,hmr2,cliff,zhang2021pymaf,pare,pixie}, usually detect individuals, segment them from the image, then use a neural network to predict 3D pose and shape in terms of the parameters of a body model like SMPL \cite{smpl}. 
\CR{Other approaches \cite{sun2021monocular,bev,qiu2023psvt} regress poses of all individuals by analyzing the full image. However, these processes lack a comprehensive understanding of the scene, failing to fully consider the interactions between humans and their environment, as well as their intentions. 
}
Methods for text-driven pose generation have also progressed rapidly \cite{posescript,jiang2023motiongpt} but the text instructions are typically ``explicit," precisely describing the pose with words.  

Thus, existing specialized systems for 3D pose estimation and generation are constrained to narrow tasks.
This is in contrast to the general-purpose reasoning exhibited by large language models (LLMs). 
Existing multimodal LLMs~\cite{llava,wu2023next,lai2023lisa,gpt4} demonstrate proficiency in perceiving and interpreting information from images and reasoning based on a wealth of world knowledge. 
They are particularly adept at describing scenes, including the appearance of people, their activities, and high-level behaviors. 
If the LLM could relate this generic world knowledge to 3D human pose and motion, it would have powerful reasoning capabilities beyond existing solutions.
That is, the LLM could bring to bear all that it has learned from both images and language for a richer and more nuanced understanding of human pose. 
Existing LLMs, however, have not yet demonstrated the ability to interpret 3D human pose.

Our hypothesis is that, long term, general purpose multimodal LLMs will subsume special-purpose methods.
Estimating 3D pose from a 2D image is fundamentally ambiguous and must use prior information or contextual cues.
Generating pose or motion from language, likewise, is ambiguous and open to interpretation.
By formulating these problems in the context of LLMs, the solutions can theoretically benefit from the LLM's broad general knowledge. 
The solutions can also benefit from interaction with a user through a language interface.
For our hypothesis to be true, LLMs must be able to understand and interpret 3D human pose.
What do they already understand about 3D pose and how can we teach them about 3D human pose?

To investigate these questions, we introduce \methodname, an approach that finetunes multimodal Large Language Models 
for predicting human pose, represented as SMPL~\cite{smpl} pose parameters. 
Our method embeds SMPL poses as a unique \texttt{<POSE>} token, prompting the LLM to output these when queried about SMPL pose-related questions. 
We extract the language embedding from this token, and use an MLP (multi-layer perceptron) to directly predict the SMPL pose parameters. 
This enables the model to take either text or images as input and subsequently output 3D body poses, as shown in Fig.~\ref{fig:teaser}. 
We maintain the vision components in a frozen state while training the SMPL projection layers and fine-tuning the LLM models with LoRA \cite{hu2021lora}. Our training strategy involves constructing question and answer pairs derived from image-to-SMPL and text-to-SMPL pose pairings, originating from pose estimation and text-driven pose generation tasks. Additionally, we integrate general multi-modal instruction-following data throughout the end-to-end training process of our model. %

We evaluate \methodname on a variety of diverse tasks, including the traditional task of 3D human pose estimation from a single image and pose generation from text descriptions.
While the metric accuracy on these classical tasks does not yet match that of specialized methods, we see this as a first proof of concept.
More importantly, 
once the LLMs are able to understand SMPL poses, they can utilize their inherent world knowledge to relate to, and reason about, human poses without the need for extensive additional data or training.  \CR{For example, as demonstrated by the right example in Fig.~\ref{fig:teaser}, \model is capable of inferring the body pose following the action depicted in the image.}
This capability gives rise to two innovative tasks concerning human poses: 
(1) \textbf{Speculative Pose Generation (SPG)}:  
In contrast to methods that generate poses like ``sitting" based on text like ``the person is sitting," in SPG we ask the LLM to speculate, for example, about ``how would the person's pose change if they were tired?"  
Such data is not in classic pose training datasets and requires an understanding of (i) what being tired does to a body and (ii) how this translates into 3D pose.
This is a significantly harder task than is considered by prior work.
(2) \textbf{Reasoning-based Pose Estimation (RPE)}: 
Contrary to conventional approaches in pose regression, our methodology does not involve providing the multimodal LLM with a cropped bounding box surrounding the individual. Instead, the model is exposed to the entire scene, enabling us to formulate queries regarding the individuals and their respective poses within that context.
For example, ``what are the poses of all the people wearing glasses?"
This requires an integration of scene understanding with 3D human pose that does not exist in current human pose regression systems.
To successfully address these tasks, the model needs two primary capabilities: 1) the ability to reason through complex and implicit text queries, integrating them with image data when available; 2) the ability to generate SMPL pose parameters based on its understanding of high-level concepts.

In summary, for the first time, we demonstrate the ability of a large vision-language model to reason about 3D human pose from images or text and to connect this with 3D SMPL parameters.
Our key contributions are as follows:
    (1) We present \textit{\model}, a multimodal Large Language Model (LLM) \CR{that can directly generate} SMPL poses. This enables the generation and estimation of human poses through reasoning from text or images.
    (2) We introduce two innovative tasks: speculative pose generation and reasoning-based pose estimation. These tasks necessitate an accurate understanding of human poses and the ability to reason using world knowledge. We have also established new benchmarks that can drive research on this topic.
    (3) Our model, \textit{\model}, demonstrates superior performance compared with other multimodal LLM baselines on the tasks of pose generation and estimation. 

%% file: sections/2_related_work.tex
\section{Related Work}
\label{sec:related_work}

Our work spans multiple research areas.
Consequently, we briefly review 3D human pose estimation from images, language and pose, and large language models. 

\vspace{+1mm}
\noindent\textbf{Human Pose Estimation.}
Human pose estimation in 2D, 3D, or over time, has a long history, which we do not review here.
Instead, we focus on work that estimates the pose of a 3D parametric body model from a single image.
Here we use the SMPL model \cite{smpl}, which produces a 3D triangulated mesh given relative body part rotations and body shape (though we ignore shape here).
SMPL is widely used, in part because it is compatible with graphics engines and because there is a large amount of training data available in SMPL format.
SMPL parameters are typically estimated from an image using one of two techniques. 
Optimization-based approaches solve for the parameters such that, when the model's 3D joints are projected into the image, they match detected 2D keypoints, subject to various priors \cite{smplify,Pavlakos2019_smplifyx,eft,Feng2023DELTA}. 
Regression-based approaches~\cite{hmr,spin,hmr2,cliff,zhang2021pymaf,pare} directly infer the pose parameters from a cropped image. 
When provided with a full image, these methods typically first detect each person in the image and then apply the regression network to tight crops.
The best regression methods are now quite accurate and robust except when there is significant occlusion, poor image quality, or unusual poses. 
\CR{
Additionally, there are methods designed for multi-person pose estimation \cite{sun2021monocular,bev,qiu2023psvt}, which are capable of directly generating body meshes for multiple people within a single image. } 
The above methods, however, do not ``understand" the semantics of human pose or relate pose to language.

\vspace{+1mm}
\noindent\textbf{Language and Human Pose.}
Given a textual description of a person's attributes, advanced image generation methods like Stable Diffusion~\cite{rombach2021highresolution} and DALL·E 2 \cite{ramesh2022hierarchical}
generate realistic 2D images of people.
These can further be conditioned on information like 2D human pose \cite{zhang2023adding}.
Such methods clearly understand properties of the human body and human pose but they output pixels and not 3D representations. 
Recent language-to-3D generation methods~\cite{poole2022dreamfusion,cao2023dreamavatar,hong2022avatarclip,liao2023tada,zhang2024teca} create 3D human shapes from textual descriptions. Yet, these methods struggle to represent  complex body poses. 
Other approaches exist that can take text input and directly produce parameters of a parametric body model like SMPL. For example, BodyTalk~\cite{bodytalk} takes human shape attributes (such as ``broad shoulders" or ``skinny") and outputs SMPL shape parameters. Similarly, \cite{briq2021towards} employs text annotations to describe a person's general action and the surrounding scene, which it uses to generate SMPL pose parameters. 
PoseScript~\cite{posescript} creates SMPL pose parameters from fine-grained textual descriptions of 3D human poses. 
While these methods are effective when test descriptions closely match the word distributions of their training data, they often lack the capability to understand or reason based on complex textual inputs. For example, PoseScript's training data lacks descriptions that relate human poses with scenes.
Since our method leverages LLMs, it can deal with more complex text queries even when trained only with the same text-to-SMPL pose pairs as PoseScript. 

Unlike all the task-specific approaches to pose estimation, action recognition, and pose generation, we develop a single,  unified, model capable of reasoning about 3D humans from images, text, or both by leveraging its general knowledge of the visual world.
Additionally, it can interact with users through conversations, discussing human poses and providing relevant responses.

\vspace{+1mm}
\noindent\textbf{Multimodal Large Language Models.}
Large Language Models (LLMs) are rapidly changing multiple fields. 
While the most powerful models like OpenAI's ChatGPT \cite{chatgpt} and GPT-4 \cite{gpt4} are private, a range of open-source LLMs such as Vicuna \cite{vicuna}, LLaMA \cite{touvron2023llama}, and Alpaca \cite{alpaca} enable research like ours.
In particular, we exploit the ability to finetune LLMs on multimodal tasks. There are two primary ways to do this.
The first leverages LLMs for decision-making guidance. Research such as \citep{yang2023mm,shen2023hugginggpt,liu2023internchat,yang2023gpt4tools,wang2023visionllm,detgpt,2023audiogpt} typically employs prompt engineering or instruction tuning. In this approach, LLMs connect separate modules via API calls. 
The LLM generates API calls to solve tasks and retrieve results.
Such an approach falls short of achieving a comprehensive understanding of new modalities. 

An alternative approach maps  modality-specific information into the language embedding space of the LLM. 
The visual modality has been a major focus in this area. Recent initiatives like LLaVA \cite{llava,liu2023improvedllava} and MiniGPT-4~\citep{zhu2023minigpt} incorporate vision encoders to interpret images and use projection layers that align image features with language embeddings. Work like LISA~\cite{lai2023lisa} generates visual information in the output, processing both images and questions to yield text and masks. 
In addition to images, MM-LLMs (Multi-Modal Large Language Models) are rapidly being developed for video \cite{2023videochat, 2023videollama} and audio \cite{2023speechgpt}. Notably, models such as PandaGPT \cite{2023pandagpt}, ImageBind \cite{girdhar2023imagebind}, and NeXT-GPT \cite{wu2023next} demonstrate the capability to handle a wide array of modalities, including text, image, audio, and video. Specifically, NeXT-GPT aligns embeddings from these four modalities with language, both as input and output. 

In this work, we investigate 3D body pose as a new modality for LLMs to process.
We explore (1) the ability of LLMs to generate  3D pose from text or image input, and (2) whether LLMs can comprehend 3D body poses and integrate this understanding into their overall functionality. 
To our knowledge, this has not previously been explored.

%% file: sections/3_method.tex
\section{Method}
\label{sec:method} 

\begin{figure*}[t]
  \centering  
  \includegraphics[width=0.99\textwidth]{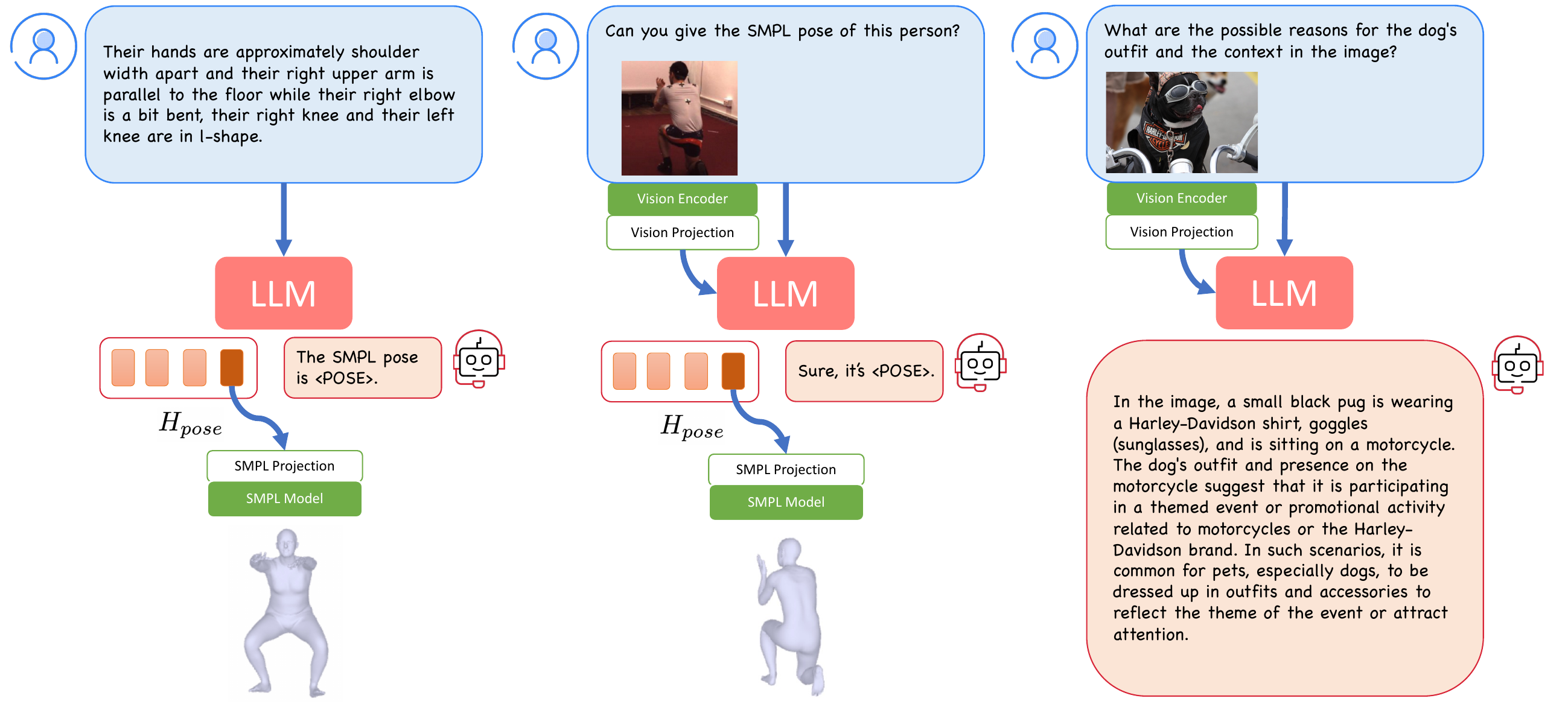}
  \vspace{-1mm}
  \caption{\small Method and Training Overview. Our model is composed of a multi-modal LLM (with vision encoder, vision projection layer and LLM), a SMPL projection layer, and the parametric human body model, i.e.~SMPL~\cite{smpl}. The multi-modal LLM processes text and image inputs (if provided) to generate textual responses. In the training phase, we focus on training the SMPL projection layer and fine-tuning the LLM, while keeping the other components frozen. The three data types used for the end-to-end training are: text-to-3D pose generation, image-to-pose estimation, and multi-modal instruction-following data. When an image is available, its information is used by the LLM to deduce an answer. If the user inquires about a SMPL pose, the LLM responds with a \texttt{<pose>} token. The embedding related to this token is then used to predict the SMPL pose parameters, leading to the generation of a body mesh, as visualized.} 
  \vspace{-3mm}
  \label{fig:method}
\end{figure*}

Our goal is to enable Large Language Models (LLMs) to comprehend human poses, represented as SMPL~\cite{smpl} pose parameters in our case. Drawing inspiration from recent advancements in multi-modal LLMs~\cite{lai2023lisa,wu2023next,golkar2023xval,ye2023natural}, we approach human pose as a distinct modality. In this framework, the LLM generates a unique token representing this modality, which is subsequently mapped to SMPL pose parameters via an MLP projection layer\footnote{Readers familiar with the geometric ``projection" of SMPL into images should not confuse that with the use of projection in this context, which effectively means  ``aligning" one representation with another.}. Leveraging the SMPL parametric model~\cite{smpl}, we can then decode this information into a three-dimensional body mesh. 
Here we describe the architecture and training strategy that integrates SMPL pose as a modality within LLMs. Once the LLM grasps the concept of 3D body pose, it gains the dual ability to generate human poses and to comprehend the world, enabling it to reason through complex verbal and visual inputs and subsequently generate human poses. 
This leads us to introduce novel tasks that are made possible by this capability, along with benchmarks to assess performance.  %

\subsection{Architecture}
The architecture of \model is illustrated in Fig.~\ref{fig:method}.  
Our approach takes text or images (if provided) as input and produces textual output. Also, when users request human pose information, it also returns the corresponding SMPL pose. 
Our model consists of a multi-modal LLM model, $f_{\phi}$, an embedding projection layer, $g_{\Theta}$, and a parametric human body model, SMPL~\cite{smpl}, represented by pose and shape parameters $\theta$ and $\beta$, respectively.
Here, we assume the $\beta$ values are all zero, corresponding to the average body shape.
Given a text string $X_q$ and an image $X_v$ as input, the model produces a textual response $Y_t = f_{\phi}(X_q, X_v)$ or $Y_t = f_{\phi}(X_q)$ in the absence of an image. 
The language embedding corresponding to $Y_t$ is represented as $H_t$. 
If \texttt{<POSE>} is present in the textual output $Y_t$, its corresponding embedding $H_{pose}$ is retrieved from $H_t$. 
The pose embedding, processed by the SMPL projection layer $g_{\Theta}$, yields the SMPL pose parameters $\theta = g_{\Theta}(H_{pose})$. 
The 3D vertices and triangles of the body mesh are then determined using the standard  SMPL function $M(\theta,\beta)$ (see \cite{smpl}).

\subsection{Training}
We keep both the vision encoder and vision projection frozen and trainf the SMPL pose projection layer $g_\Theta$.  Additionally, we employ LoRA~\cite{hu2021lora} to finetune the LLM, with its parameters denoted as $\phi_{lora}$. The final set of optimizable parameters is $\{\phi_{lora}, \Theta\}$. 
With the provided ground truth textual output $\hat{Y}_t$ and SMPL pose parameters $\hat{\theta}$, we optimize the model using the following objective function: 
\begin{equation}
    \mathcal{L} = \lambda_{t}\mathbf{CE}(\hat{Y}_t, Y_t) + \lambda_{\theta}|\hat{\theta} - \theta|.
\label{eq:loss}
\end{equation} 
\CR{The first term is the cross-entropy loss}, while 
the second, pose loss, is the L1 difference between the ground truth and estimated pose parameters. 
$\lambda_t$ and $\lambda_{\theta}$ serve as the weights for their respective loss terms. 
To train our multi-modal LLM model, we construct data by leveraging existing task-specific datasets below. 

\noindent\textbf{Text to Pose Generation.} 
A 3D human pose can be generated from a detailed textual description of the pose. The data pairs in this case are SMPL pose parameters and detailed text description labels  $\{X_q, \hat{\theta}\}$. 
To fit this data into a question-answer format, we employ templates such as  ``\texttt{{USER}}: \texttt{\{description\}}, \texttt{\small can you give the SMPL pose of this person.}  \texttt{{ASSISTANT}}: \texttt{\small Sure, it is} \texttt{<POSE>}.", where \texttt{\{description\}} contains the pose descriptions $X_q$ from the dataset.  

\noindent\textbf{Human Pose Estimation.}  
Conventional methods of 3D human pose estimation~\cite{hmr2,spin} typically involve using cropped images to regress SMPL body shape and pose parameters. 
Similarly, we use pairs of cropped images and SMPL pose parameters $\{X_v, \hat{\theta}\}$. 
To format the data suitably for visual question answering, similar to text to pose generation, we use a question-answer template like ``\texttt{{USER}}: \texttt{<IMAGE>} \texttt{Can you provide the SMPL pose of the person in the center of this image?} \texttt{{ASSISTANT}}: \texttt{\small Sure, the SMPL pose of this person is} \texttt{<POSE>}.", where \texttt{<IMAGE>} is a placeholder for the input image tokens.  The corresponding ground truth SMPL pose parameters $\hat{\theta}$ are used to calculate the pose loss as in Equation~\ref{eq:loss}. During training, we also use other templates to generate question-answer data to ensure  diversity; please see \supmat for details. 

\noindent\textbf{Multi-Modal Instruction-following.} 
In order to maintain the multi-modal LLM's inherent capability for multi-turn conversations, we use a multi-modal instruction-following dataset during training. Following LLaVA-V1.5~\cite{liu2023improvedllava}, we utilize the LLaVA-V1.5-MIX665K\footnote{\href{https://huggingface.co/datasets/liuhaotian/LLaVA-Instruct-150K}{\tt liuhaotian/LLaVA-Instruct-150K}} dataset, which is created through queries made to GPT-4.

\subsection{Reasoning about Human Pose} 
After training, our model is capable of estimating SMPL poses from single images, generating poses based on detailed descriptions, and facilitating question-and-answer conversations. 
Remarkably, even without integrating SMPL pose into multi-turn conversations or linking complex phrases with SMPL pose, our model demonstrates a {\em zero-shot}  capability for reasoning about human poses within multi-turn dialogues. This suggests that the model is able to interweave reasoning and world knowledge with the SMPL pose representation. 
Therefore, in addition to conventional evaluation approaches for human pose and generation tasks, we introduce two new tasks that require reasoning skills: Speculative Pose Generation and Reasoning-based Pose Estimation. These new tasks leverage the model's ability to apply reasoning in the context of human pose analysis.

\noindent\textbf{Speculative Pose Generation (SPG).}  
In this task, rather than using explicit pose descriptions from the text-to-pose generation dataset, users pose indirect questions about a person's state, requiring the LLM to deduce and generate the appropriate pose. For instance, a user might ask, ``\texttt{{USER}}: \texttt{\{descriptions\_implicit\}}, \texttt{\small can you give the SMPL pose of this person?}  \texttt{{ASSISTANT}}: \texttt{\small Sure, it is} \texttt{<POSE>}."
Here, \texttt{\{description\_implicit\}} represents speculative queries such as ``This man is proposing marriage, what pose might he be in?". This kind of inquiry requires an understanding of global concepts such as ``marriage" and the capacity to logically deduce the individual's pose, followed by the generation of SMPL pose parameters. 
 To create an evaluation dataset, we use pose descriptions from the PoseScript~\cite{posescript} dataset as a source. We then query GPT4 to reformulate these descriptions into questions about the activities associated with each pose, generating a total of 20k responses, of which 780 examples are used for evaluation. These responses are then manually reviewed and corrected as needed. 

\noindent\textbf{Reasoning-based Pose Estimation (RPE).} 
Standard human pose estimation methods typically 
first run a person detector and then only process a cropped image around the person.
This ignores scene context, which can be useful in reasoning about human pose. 
In contrast, RPE lets users make inquiries about an image before requesting details about a person's pose. Specifically, we define RPE as: 
``\texttt{{USER}}:\texttt{<IMAGE>} \texttt{\{description\_person\}}, \texttt{\small can you give the SMPL pose of this person?}  \texttt{{ASSISTANT}}: \texttt{\small Sure, it is} \texttt{<POSE>}."  In this case, \texttt{\{description\_person\}} could be queries about a particular individual, such as ``The man with black hair", or ``the woman near the stairs". The model is required to interpret the scene context and generate the SMPL pose parameters for the individual fitting the description. 
To evaluate this task, we start with image-to-SMPL pose pairs from standard pose estimation evaluation datasets. We then use GPT4V to generate descriptions of the individuals in these images. The generated descriptions are subsequently refined manually. Specifically, we sample 50 multiple-person images from the 3DPW~\cite{vonMarcard2018} test set. 
For each individual, we collect descriptions that cover \texttt{{behavior, outfits, pose, shape, summary}}, where \texttt{summary} summarizes all the other attributes. This process leads to a total compilation of 250 question and answer pairs for evaluation.  
For more details of the collection pipeline, please see the \supmat

%% file: sections/4_experiment.tex
\section{Experiments}
\label{sec:experiment}
\input{tables/01_reason_generation_sota}
\begin{figure}[t]
  \centering  
  \includegraphics[width=0.5\textwidth]{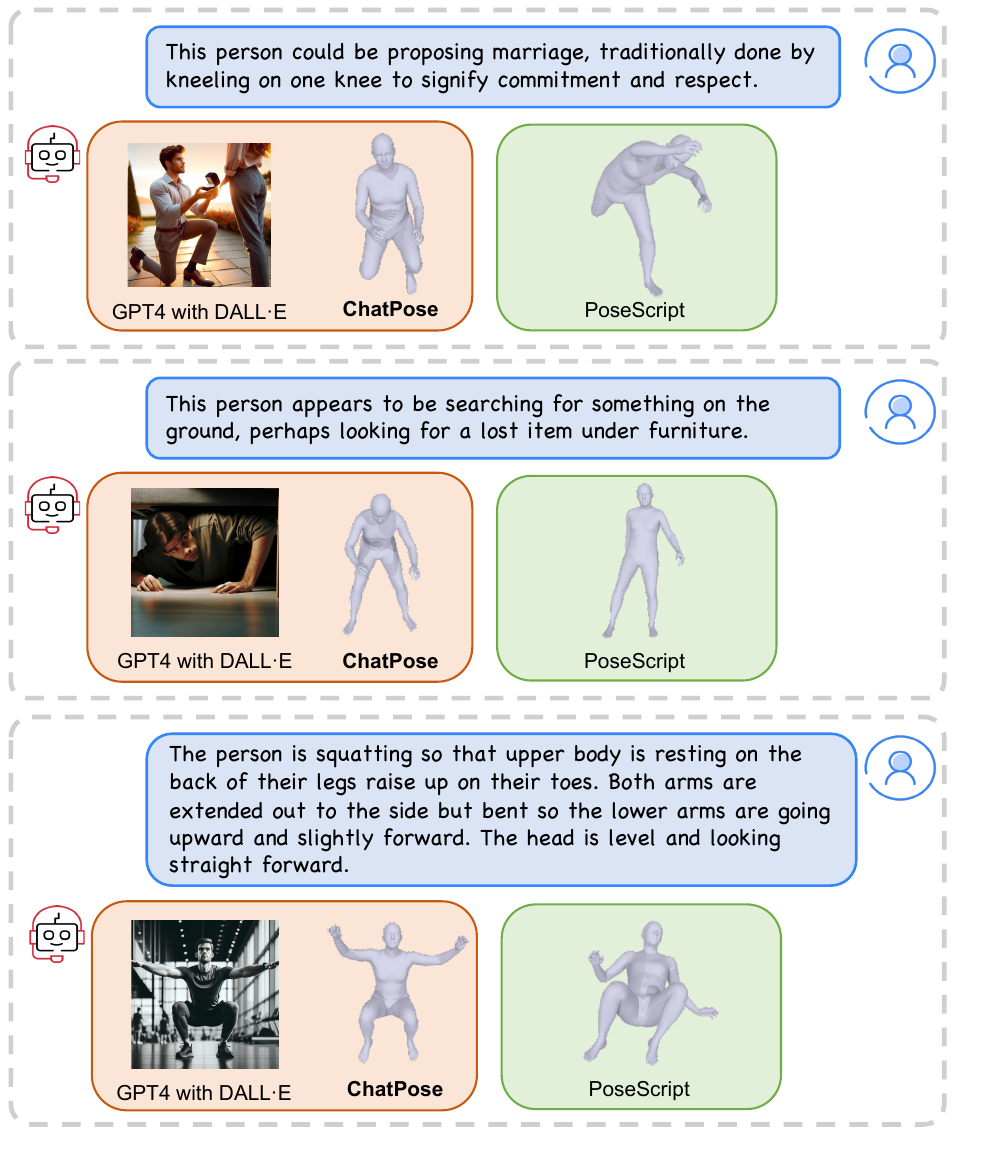}
  \vspace{-7mm}
  \caption{\textbf{Pose Generation.} GPT-4 (DALL·E)~\cite{gpt4} generates images that depict the correct pose but does not explictly generate 3D poses. In contrast, PoseScript~\cite{posescript} is a task-specific method for 3D pose from language but it is not able to relate high-level concepts like ``searching under furniture" with 3D pose. In contrast, \model, understands high-level concepts and how to relate them to 3D pose.
  The methods in orange address SPG, while the green region indicates the ``classical" approach. 
  The first two query examples are sourced from our SPG benchmark, which offers implicit text queries regarding human poses. The third example is derived from the PoseScript test set, which has detailed descriptions of human poses. 
  }
  \vspace{-4mm}
  \label{fig:exp_pose_generation} 
\end{figure}

\begin{figure*}[h]
  \centerline{      \includegraphics[width=0.99\textwidth]{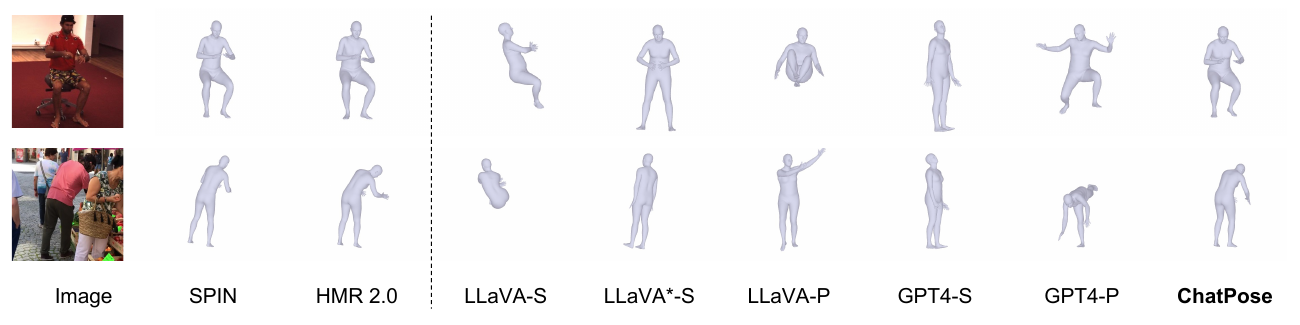}}
  \vspace{-0.15in}
  \caption{\small  
  We compare multi-modal LLMs (LLaVA~\cite{llava}, GPT-4~\cite{gpt4}) and traditional HMR-style methods (HMR2.0~\cite{hmr2}, SPIN~\cite{spin}) for \textbf{classical human pose estimation}. \llava\/* is LLaVA  fine-tuned with keypoint data. 
  }
  \label{fig:exp_pose_estimation}
  
\end{figure*}

\begin{figure}[h]
\centerline{      \includegraphics[width=0.5\textwidth]{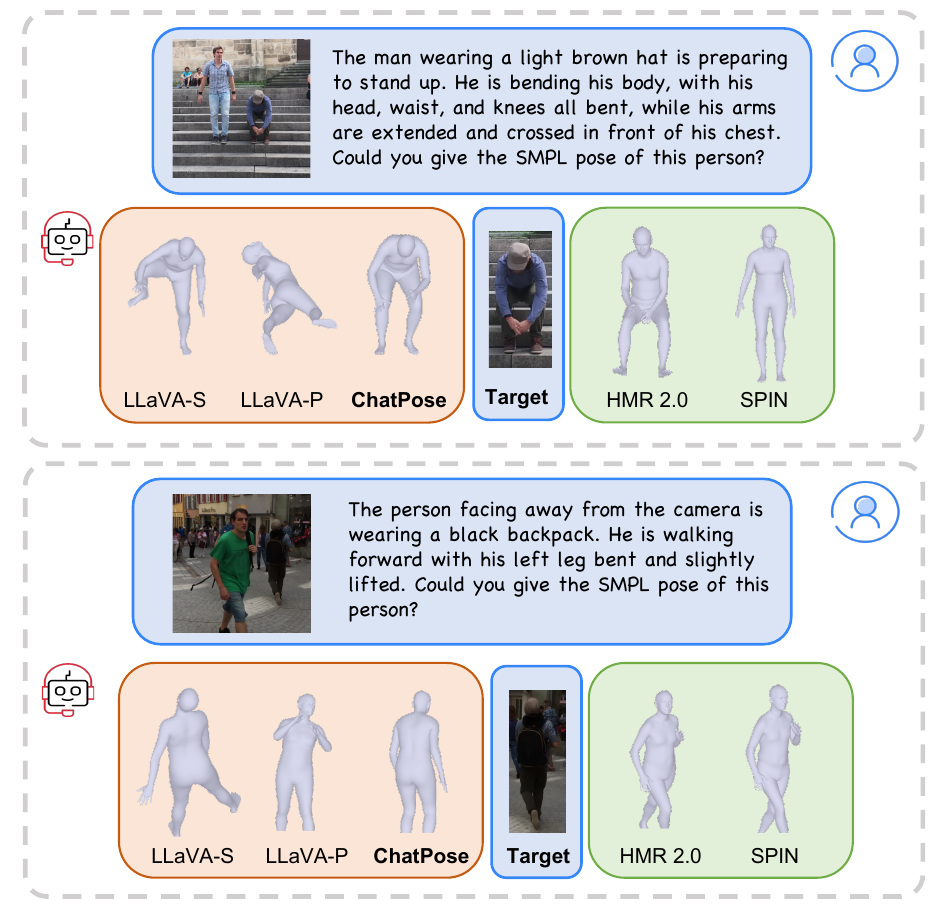}}
  \vspace{-0.1in}
  \caption{\small 
  Comparison with LLaVA~\cite{llava} and classical HMR-style  methods (HMR2.0~\cite{hmr2} and  SPIN~\cite{spin}) on \textbf{reasoning-based human pose estimation}. 
 For each method, we utilize the entire image provided by the user as input, without applying cropping.  
 Methods involving LLMs are highlighted in orange, while those that are purely task-specific methods, are marked in green.
 }
  \vspace{-6mm}

  \label{fig:exp_reasoning_hmr}
\end{figure}

\input{tables/02_reason_hmr_sota}
\input{tables/03_hmr_sota}

We employ LLaVA-1.5V-13B~\cite{llava} as the multimodal LLM backbone, with CLIP~\cite{clip} for vision encoding and Vicuna-13B~\cite{zheng2023judging}, finetuned from Llama 2~\cite{touvron2023llama} on conversational data, for the LLM backbone. We maintain the CLIP encoder and vision projection layer, while training the SMPL projection layer from scratch and fine-tuning the LLM using LoRA.
The SMPL projection layer is an MLP with layer dimensions of [5120, 5120, 144]. Following previous work~\cite{hmr2, spin}, our network predicts 6D rotations~\cite{zhou2019continuity} for the SMPL pose, which are converted into rotation matrices for loss computation. 
For further implementation details, training details, our ablation study, and details about LLM backbones, please see \supmat

\subsection{Datasets}
\noindent\textbf{Text to Pose Generation.}  
We use the text-to-SMPL pose pairs from PoseScript~\cite{posescript}, which features textual descriptions of 20k diverse human poses derived from the AMASS \cite{amass} dataset. 
Within this dataset, 6.5k texts are human-annotated and there are six types of automated labels for the entire set of 20k poses. 
Our training employs their designated training set of approximately 14k pairs. Additionally, we observe that the automatically generated labels in the dataset exhibit significant noise. Thus, we prioritize human labels when available; in their absence, we randomly select one of the automated labels for each pose. 

\noindent\textbf{Human Pose Estimation.} 
In line with prior research on ``classical" 3D human pose and shape regression, we employ datasets from Human3.6M~\cite{human36m}, MPI-INF-3DHP~\cite{mehta2017monocular}, COCO~\cite{lin2014microsoft}, and the MPII dataset~\cite{mpii} for training. 
These datasets include training pairs of images with ground-truth or pseudo-ground-truth SMPL pose parameters. 
Note that we ignore the SMPL shape parameters here. 
Unlike previous methods, which typically use significant data augmentation (e.g.~\cite{pare,lin2023osx}), our approach solely uses tightly cropped images without any additional augmentation such as blur or occlusion. 
Despite this, our model still demonstrates good generalization to these scenarios, suggesting that the network is able to leverage its general visual capabilities.

\subsection{Evaluation Metrics and Baselines} 
\noindent\textbf{Generation.} For both the standard text-to-pose generation task and our new  speculative pose generation (SPG) task, we use the evaluation metrics established in PoseScript~\cite{posescript}. 
We report the text-to-pose recall rate $R^{T2P}$ and the pose-to-text recall rate $R^{P2T}$ of the retrieval models trained on real poses and evaluated on generated poses. 
\CR{Following previous work~\cite{posescript}, for the SPG task, the retrieval model is retrained for evaluation using SPG training data.}

\noindent\textbf{Estimation.} 
To evaluate traditional and reasoning-based 3D pose estimation,  
we use the traditional metrics: Mean Per-Joint Position Error (MPJPE) and this error after rigidly aligning the posed body with the ground truth (PA-MPJPE). Additionally, we introduce the Mean Per-Joint Rotation Error (MPJRE) to more directly evaluate body pose accuracy.
To evaluate human pose estimation, we select 200 samples from the 3DPW~\cite{vonMarcard2018} and Human3.6M~\cite{human36m} test sets. 
\CR{
To assess \model's performance on SPG and RPE tasks, we introduce several baseline methods:
\begin{itemize}
  \item \noindent\textbf{LLaVA*}. Instead of utilizing the pose token \texttt{<POSE>}, human poses can be represented through language, such as textual descriptions of keypoint locations. Using the same dataset pairs as in \model, we formulate VQA pairs as described in \supmat for training. We then fine-tune the base model LLaVA, referred to as \llava*, with results shown in Table \ref{tab:exp_hmr} and Fig.~\ref{fig:exp_pose_estimation}. 
  \item \noindent\textbf{LLaVA-S, LLaVA*-S, and GPT4-S}. For the RPE task, we initially request LLMs, such as LLaVA~\cite{llava}, LLaVA*, and GPT4~\cite{gpt4}, to provide textual descriptions of the keypoint locations for the target individual, and then apply SMPLify~\cite{smplify} to optimize the human poses based on these keypoint locations.
  \item \noindent\textbf{LLaVA-P and GPT4-P}. Similarly, for RPE and SPG tasks, we use LLMs like LLaVA~\cite{llava} and GPT4~\cite{gpt4} to describe human poses in response to questions, and then generate SMPL poses with PoseScript~\cite{posescript} from these descriptions. We show the RPE results in Figure~\ref{fig:exp_reasoning_hmr} and SPG comparison in \supmat
\end{itemize} 
}

\subsection{Pose Generation} 
We evaluate \model's pose generation capabilities on both the classical task and the new SPG task.
Figure~\ref{fig:exp_pose_generation} shows how \model handles detailed and speculative queries, outperforming PoseScript in complex scenarios involving reasoning. 
While \model and DALL·E produce different output modalities (3D poses vs images), they both ``understand" the concepts.  
Quantitatively, as Table~\ref{tab:exp_pose_generation} shows, \model performs comparably to PoseScript on classical tasks (with detailed pose descriptions) and outperforms it on speculative pose generation. 

\subsection{Pose Estimation} 
Figure~\ref{fig:exp_pose_estimation} and Table~\ref{tab:exp_hmr} show  qualitative and quantitative results on classical human pose estimation. \model outperforms other Multi-modal LLMs, yet it does not match the performance of methods designed and trained specifically to estimate 3D human pose. 
This is not surprising and we see these results as a first proof-of-concept.
For failure cases, please see the \supmat 
In reasoning-based human pose estimation, 
\model outperforms both task-specific and multi-modal LLM methods. This  is illustrated in Fig.~\ref{fig:exp_reasoning_hmr} and Table~\ref{tab:exp_reason_hmr}. 
Notably, the MPJPE is heavily affected by the global orientation, while PA-MPJPE lessens this impact, offering a truer reflection of body pose accuracy. 
\model has trouble estimating global orientation of the person; this could likely be addressed by additional training. 

We also found that \model generalizes well to strong occlusions. 
Even without any data augmentation during training.
This suggests that it is able to leverage its general visual knowledge about occlusion in solving the human pose estimation problem.
See \supmat for examples.

\subsection{GPT-Assisted Evaluation}
When training \model to understand 3D pose, it is critical that it does not forget its general knowledge. To evaluate this, we follow LLaVA's~\cite{llava}, using GPT4-Assisted evaluation. 
Table~\ref{tab:exp_eval_gpt4} shows \model slightly lags LLaVA, indicating \model successfully combines 3D pose abilities with its vision and language understanding.

\input{tables/eval_gpt4} 

\subsection{Ablation study} 
\label{ablation}
We evaluate the impact of different aspects of \model, including human pose representations, multi-modal LLM backbones, and various datasets. Please refer to \supmat.

%% file: tables/01_reason_generation_sota.tex
\begin{table}[t]
	\centering
	\scriptsize
	\resizebox{\columnwidth}{!}{
		\begin{tabular}{|l|c|c|c|c|}
			\toprule
			& \multicolumn{2}{c|}{ \posescript \cite{posescript}} & \multicolumn{2}{c|}{SPG Benchmark}\\
			\cmidrule(lr){2-5}
			\textbf{Method}  & {$R^{P2T}$} $\uparrow$ & {$R^{T2P}$} $\uparrow$  & {$R^{P2T}$} 
			$\uparrow$ & {$R^{T2P}$} $\uparrow$ \\
			\midrule
                \posescript \cite{posescript}     &     \textbf{22.6}~/~\textbf{31.0}~/~\textbf{42.3}      &     22.4~/~32.1~/~43.6    &      1.9~/~3.8~/~6.5    &      2.8~/~4.3~/~7.2~     \\
                \model       &     17.6~/~25.3~/~35.8     & \textbf{28.0}~/~\textbf{39.0}~/~\textbf{54.4}
				&      \textbf{8.6}~/~\textbf{14.2}~/~\textbf{20.8} &      \textbf{10.9}~/~\textbf{16.9}~/~\textbf{25.3}
				\\
			\bottomrule
		\end{tabular} 
	}
 \vspace{-2mm}
	\caption{
	            \cheading{Comparison of classical and speculative pose generation} 
	            \captionArrows. 
             Top 5~/~10~/~20 retrieval recall rates are reported for pose generation on the PoseScript test set and our new SPG Benchmark. 
	}
  \vspace{-5mm}
	\label{tab:exp_pose_generation}
\end{table}

%% file: tables/02_reason_hmr_sota.tex
\begin{table*}[h]
	\centering
	\resizebox{1.0\linewidth}{!}
	{
	\setlength{\tabcolsep}{1mm}
		\begin{tabular}{l|l|c|c|c|c|c|c}
		\toprule
		\multirow{2}[4]{*}{\textbf{Stage 1}} & 
		\multirow{2}[4]{*}{\textbf{Stage 2}} &
		\multicolumn{5}{c|}{\boldmath{}Text Description\unboldmath{}} & \multirow{2}[4]{*}{Averaged} \\
		\cline{3-7}
		& & Behavior & Shape & Outfit & Pose & Summary \\
		\midrule
		SPIN \cite{spin} & - & ~244.9~/~107.3~/~12.4~ & ~244.9~/~107.3~/~12.4 & 244.9~/~107.3~/~12.4 & 244.9~/~107.3~/~12.4 & 244.9~/~107.3~/~12.4 & 244.9~/~107.3~/~12.4 \\ 
        HMR 2.0 \cite{hmr2} & - & \textbf{225.2}~/~105.7~/~12.1 & \textbf{225.2}~/~105.7~/~12.1 & \textbf{225.2}~/~105.7~/~12.1 & \textbf{225.2}~/~105.7~/~12.1 & \textbf{225.2}~/~105.7~/~12.1 & \textbf{225.2}~/~105.7~/~12.1 \\ 
		\midrule
        LLaVA \cite{llava} & SMPLify \cite{smplify} & 490.7~/~200.6~/~20.9 & 462.3~/~204.3~/~20.2 & 481.1~/~198.7~/~20.0 & 480.9~/~207.4~/~21.1 & 490.7~/~207.4~/~21.1 & 481.1~/~203.7~/~20.7 \\ 
        LLaVA \cite{llava} & PoseScript \cite{posescript} & 370.8~/~182.3~/~17.5 & 407.8~/~191.3~/~18.0 & 440.7~/~190.4~/~17.6 & 363.2~/~177.9~/~17.4 & 391.5~/~191.9~/~17.8 & 394.8~/~186.8~/~17.7 \\ 
        \model (Ours) & - & 307.9~/~\textbf{102.9}~/~\textbf{12.1} & 269.9~/~\textbf{103.7}~/~\textbf{12.0} & 265.6~/~\textbf{102.6}~/~\textbf{11.8} & 277.9~/~\textbf{96.0}~/~\textbf{11.7} & 253.6~/~\textbf{103.8}~/~\textbf{11.7} & 275.0~/~\textbf{101.8}~/~\textbf{11.9}  \\
		\bottomrule
		\end{tabular}%
		}
  \vspace{-0.1in}
		\caption{
			Comparison of \textbf{reasoning-based pose estimation} with different text descriptions. 
			MPJPE~/~PA-MPJPE/~MPJRE ($\times100$) on the RPE benchmark are reported. Examples of each description type are in the \supmat
			\captionBold. 
}
\vspace{-2mm}
\label{tab:exp_reason_hmr}

\end{table*}
    

%% file: tables/03_hmr_sota.tex
\begin{table}[t]
	\centering
	\scriptsize
	\resizebox{\columnwidth}{!}{
		\begin{tabular}{|l|c|c|c|c|c|}
			\toprule
			&  \multicolumn{3}{c|}{ \threedpw \cite{threedpw} } & \multicolumn{2}{c|}{\humanthreesix \cite{human36m}}\\
			\cmidrule(lr){2-6}
			\textbf{Method} & {MPJPE $\downarrow$} & {PA-MPJPE $\downarrow$} & {MPJRE $\downarrow$} & {MPJPE  $\downarrow$} & {PA-MPJPE $\downarrow$} \\
			\midrule
			SPIN \cite{spin}  &  102.9 &     62.9     &     10.1   &  61.9 &    42.6            \\
			HMR 2.0 \cite{hmr2}   &     91.0      &     58.4    &  9.2 &    50.0    &      33.6        \\
			\midrule
                \llava-S \cite{llava}          &     440.8    &      205.4    & 21.8  &   461.3   &      195.4      \\
                \llava*-S \cite{llava}       &     232.1 &      101.1    &      12.8  &      246.0 & 118.2   \\
                GPT4-S \cite{gpt4}      &     322.0  &      136.7    &      16.0  &      336.9 & 144.0   \\
                \llava-P \cite{llava}      &     335.2    &      172.3    &  16.5 &    334.1    &      172.5  \\
                GPT4-P \cite{gpt4}        &     396.5    &      203.4    &      18.6   &   354.1  & 203.5  \\
                \model (Ours)     &     163.6      &     81.9    &   10.4    &      126.0     & 82.4     \\
			\bottomrule
		\end{tabular} 
	}
 \vspace{-0.1in}
	\caption{
	            \cheading{Comparison on Human Pose Estimation}
	            %
				MPJPE (mm), PA-MPJPE (mm), and MPJRE ($\times100$) are reported.
	}
  \vspace{-4mm}
 
	\label{tab:exp_hmr}
\end{table}

%% file: tables/eval_gpt4.tex
\begin{table}[t]
\centering
\footnotesize	
\begin{tabular}{l|cccc}
\hline
Method & Conv & Detail & Complex & All \\
\hline
LLaVA-V1-13B~\cite{llava} & 83.1 & 75.3 & 96.5 & 85.1 \\
LLaVA-V1.5-13B~\cite{liu2023improvedllava} & 84.4 & 81.0 & 93.9 & 86.5 \\
\model (Ours) & 78.8 & 76.2 & 96.7 & 84.0 \\
\hline
\end{tabular}
\vspace{-0.1in}
\caption{\cheading{GPT4-Assisted Evaluation} 
``Conv," ``Details," and ``Complex" signify three categories of questions produced by the LLaVA data generation pipeline, covering conversation, detailed description, and complex reasoning. 
}
\label{tab:exp_eval_gpt4}
  \vspace{-5mm}
\end{table}


%% file: sections/5_conclusion.tex
\section{Conclusions}
\label{sec:conclusions} 
\model makes a first step towards integrating 3D human pose estimation with the general reasoning capabilities of LLMs.
This study teaches us several things. 
First, multimodal LLMs can be fine-tuned to infer 3D human pose from images. In particular, they are able to infer the real-valued rotations of human body parts.
To our knowledge, this is the first demonstration that such models can directly solve this task.
Second, the model can connect 3D human pose with language. This is important because it opens up many possibilities both for applications and for training.
Third, we have demonstrated new use cases in which a user can chat with the language model about 3D human pose using text and images.
We think this is the beginning of a rich space that will open up new ways of training and using LLMs to reason about 3D human pose. %

\noindent\textbf{Limitations.} 
The accuracy of our 3D pose estimation from images is below recent specialized regressors.
Better quality data relating language to pose is needed.  A key lesson of recent LLM research is that the scale and quality of the data is key. 
%
Additionally, freezing the vision encoder is a limitation which could be overcome with a more powerful backbone or by fine-tuning the whole model on more data.


\noindent\textbf{Future work.}
Future work should also improve the ability of \model to have multi-turn conversations about 3D pose.
It should also be possible to enable pose editing, cf.~\cite{delmas2023posefix}.
It should be straightforward to extend our work to infer and reason about 3D body shape and human movement. 
The extension to video input is particularly promising given recent progress on video models, which have broad knowledge about the 3D world and human behavior, e.g.~\cite{blattmann2023stable}.


%% file: sections/6_acknowledgement.tex
{\small \noindent\textbf{Acknowledgements.}
We thank Weiyang Liu, Haiwen Feng and Longhui Yu for discussions and proofreading. 
We also thank Naureen Mahmood and Nicolas Keller for support with data.
This work was partially supported by the Max Planck ETH Center for Learning Systems.  
CoI disclosure: \url{https://files.is.tue.mpg.de/black/CoI_CVPR_2024.txt}.
}


%% file: sections/X_suppl.tex
\maketitlesupplementary

\section{Training Data Details}
As described in the Method, we construct question and answer pairs to finetune a multi-modal LLM; specifically we use text-to-SMPL pose and image-to-SMPL pose pairs. 
Details of the question list are illustrated in Table \ref{tab:training_data_posescript} and Table \ref{tab:training_data_hmr}, while  example answers are shown in Table \ref{tab:training_data_answer}. 

\begin{table}[h!]\centering
\begin{minipage}{0.99\columnwidth}\vspace{0mm}    \centering
\begin{tcolorbox} 
    \centering
    \small
     \hspace{-6mm}
\begin{itemize}[leftmargin=2mm]
\setlength{\itemsep}{1pt}
\item ``\texttt{<image>} Can you predict the SMPL pose of the person in this image?" 
\item ``\texttt{<image>} There is a person in the middle of the image, please output this person's SMPL pose."
\item ``\texttt{<image>} What is the human pose in this image? Please respond with SMPL pose."
\item ``\texttt{<image>} What is the person doing in this image? Please output SMPL pose."
\item ``\texttt{<image>} There is a person in the middle of the image, use SMPL to describe the pose."
\end{itemize}
\end{tcolorbox}
\vspace{-3mm}
\caption{The list of questions for training \model with image-to-SMPL pose pairs.}
    \label{tab:training_data_hmr}
\vspace{-3mm}
\end{minipage}
\end{table}

\begin{table}[h!]\centering
\begin{minipage}{0.99\columnwidth}\vspace{0mm}    \centering
\begin{tcolorbox} 
    \centering
    \small
     \hspace{-6mm}
\begin{itemize}[leftmargin=2mm]
\setlength{\itemsep}{1pt}
\item ``The SMPL pose is \texttt{<POSE>}."
\item ``It is \texttt{<POSE>}."
\item ``The SMPL format of this person's pose is \texttt{<POSE>}."
\item ``Sure, it is \texttt{<POSE>}."
\item ``Sure, the SMPL pose is \texttt{<POSE>}."
\item ``\texttt{<POSE>}."
\item ``The SMPL pose of the person is \texttt{<POSE>}."
\item ``Sure, \texttt{<POSE>}."
\end{itemize}
\end{tcolorbox}
\vspace{-3mm}
\caption{The list of answers for training \model with SMPL pose as the output.}
    \label{tab:training_data_answer}
\end{minipage}
\vspace{-3mm}
\end{table}

\begin{table*}\centering
\begin{minipage}{1.99\columnwidth}\vspace{0mm}    \centering
\begin{tcolorbox} 
    \centering
    \small
     \hspace{-6mm}
\begin{itemize}[leftmargin=2mm]
\setlength{\itemsep}{1pt}
 \item  ``I have a word description of a person's pose, can you give the SMPL pose of this person? \texttt{\{description\}}"
 \item ``There is a person: \texttt{\{description\}} Please output this person's SMPL pose."
 \item ``\texttt{\{description\}} Give the SMPL pose."
 \item ``What's the SMPL pose of this person? \texttt{\{description\}}"
 \item ``Use SMPL pose to describe this person's behavior. \texttt{\{description\}}"
 \item ``There is a person doing this: \texttt{\{description\}} Can you use SMPL pose to describe the pose?"
 \item ``A person is described as: \texttt{\{description\}} Use the SMPL pose to reflect this."
 \item ``Human pose is described as words: \texttt{\{description\}} The SMPL pose is?"
 \item ``Human pose can be described as words: \texttt{\{description\}} And it can also be described in SMPL pose format, can you output this?"
\end{itemize}
\end{tcolorbox}
\caption{The list of questions for training \model with text-to-SMPL pose pairs. Where \texttt{\{description\}} is the text description from the dataset.}
    \label{tab:training_data_posescript}
\end{minipage}
\end{table*}

\section{Benchmark Details}
\noindent We introduce two benchmarks, speculative pose generation (SPG) and reasoning-based pose estimation (RPE), to evaluate the performance on reasoning about human poses. 

\paragraph{SPG Benchmark.} Unlike traditional text-to-pose generation tasks, speculative pose generation requires the model to reason about, and interpret, indirect pose descriptions and to generate appropriate 3D poses. Consequently, a novel benchmark for evaluation is necessary. We utilize the PoseScript dataset \cite{posescript}, which provides direct pose descriptions, as a starting point. \CR{Subsequently, we visualize the pose from four viewpoints and feed the visual result along with the direct pose description into GPT-4V~\cite{gpt4}, prompting it to generate implicit descriptions of associated activities, as shown in Figure \ref{fig:spg_benchmark}. To improve the generation quality, we design a chain-of-thought mechanism, in which we ask GPT-4V to answer four questions before generating the speculative pose descriptions.
The details of the query input are presented in Table~\ref{tab:SPG_query}.} We then manually check these labels and construct instruction data containing 780 text-pose pairs formatted as follows: ``\texttt{{USER}}: \texttt{\{descriptions\_implicit\}}, \texttt{\small can you give the SMPL pose of this person?}  \texttt{{ASSISTANT}}: \texttt{\small Sure, it is} \texttt{<POSE>}."
Here, \texttt{\{description\_implicit\}} represents the speculative queries generated by GPT4. 

\begin{figure*}[!h]
  \centering  
  \includegraphics[width=0.99\textwidth]{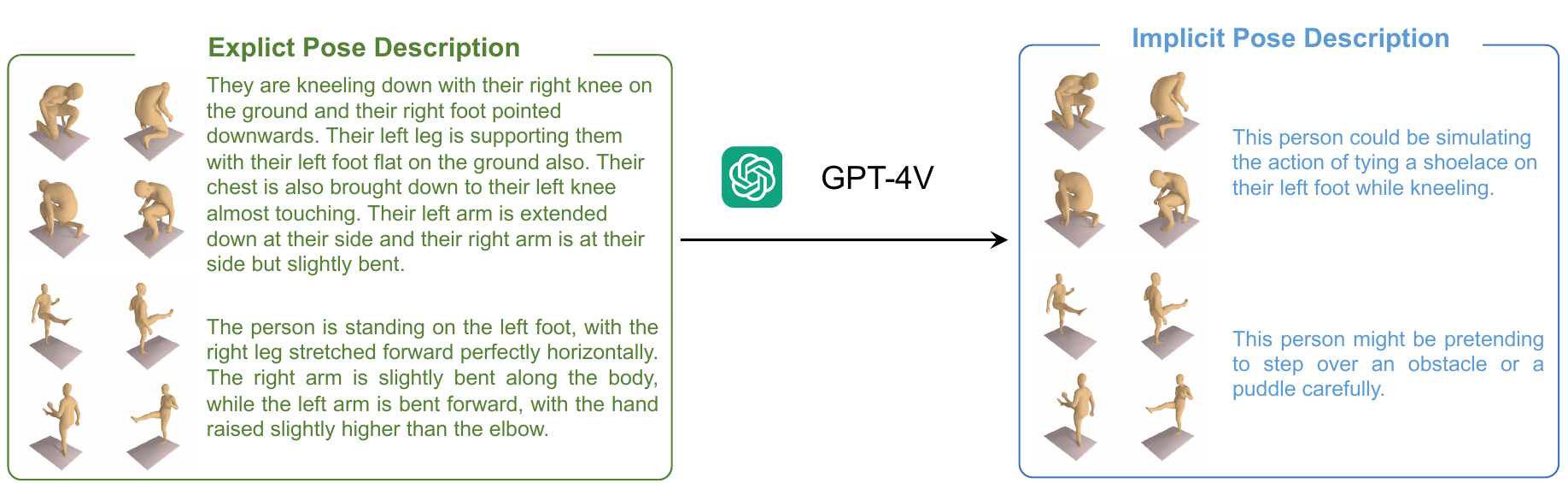}
  \caption{\small Illustration of the annotation pipeline that generates implicit pose description for our SPG benchmark. We take the fine-grained explicit pose descriptions from PoseScript \cite{posescript} and visualize the described pose from four viewpoints, and then query GPT4 to reformulate them into indirect pose descriptions.}
  \label{fig:spg_benchmark}
\end{figure*}

\begin{table*}\centering
\begin{minipage}{1.99\columnwidth}\vspace{0mm}    \centering
\begin{tcolorbox} 
    \centering
    \small
     \hspace{-6mm}
\begin{tabular}{p{0.99\columnwidth}}
 \begin{minipage}{0.99\columnwidth}\vspace{0mm}
As an AI visual assistant specializing in human pose analysis, you will receive a visual depiction of a person, captured from multiple views, and a detailed, fine-grained textual description of his/her pose. \\
Your task is to infer the possible daily activities, ball games, and other behaviors that the person is mimicking. Offer a high-level interpretation without delving into the minutiae of joint positions. Concentrate on high-level descriptions of daily activities, ball games, and behaviors evident from the visual and textual information provided. If the pose resembles a specific yoga pose, be sure to mention the name of the yoga pose. \\

Ensure that your answers are clear enough to allow users to accurately mimic and replicate the pose based on your description. Avoid overly vague and ambiguous descriptions such as "This person is doing a balancing behavior" or "The person is warming up". Your answer should be as diverse as possible and minimize the use of terms like "balance", "stretch", "warm-up", and "flexibility". \\ 

Prior to formulating the pose description, think and answer the following questions: \\ 

1. Which yoga pose the person might be doing? What are the differences between the visualized pose and standard yoga pose? \\ 
2. What everyday activity might the individual be engaging in? \\ 
3. Which sporting activity appears to be mimicked by the individual?  \\
4. Could there be other actions the person is undertaking? \\ 

Based on your responses to the above questions, craft 5 responses describing the pose, each starting with "\{number\}. This person," accompanied by a succinct one or two sentences.
Example answers and pose descriptions: \\ 

Answer to the questions: \\
1. The individual seems to be adopting a yoga pose, resembling the "Natarajasana" or "Lord of the Dance Pose." \\
2. The individual could be reaching for an item on a high shelf. \\
3. It appears the individual is imitating a basketball player. \\
4. Additionally, the person might be engaging in an activity such as watching a movie with a friend. \\ 

Pose descriptions: \\
1. This person is executing the "Downward-Facing Dog" yoga pose. \\
2. This person is making a marriage proposal. \\
3. This person is kneeling on one knee, potentially in a protest. \\
4. This person is participating in basketball, performing a jump shot. \\
5. This person seems to be looking for something on the ground. 

\end{minipage}
    \end{tabular}
    
\end{tcolorbox}
\vspace{-2mm}
\caption{Example to query GPT4 for implicit pose descriptions.}
    \label{tab:SPG_query}
\end{minipage}
\vspace{-4mm}
\end{table*}

\paragraph{RPE Benchmark.} To establish the reasoning-based pose estimation benchmark, we begin by selecting 50 multiple-person images from the 3DPW \cite{threedpw} test set. Subsequently, we employ GPT4V to generate descriptions of the individuals depicted in these images, covering attributes like \texttt{behavior, outfits, pose, shape, summary}, with \texttt{summary} summarizing all the other attributes. Notably, during our experiments, we observe that GPT4V~\cite{gpt4} consistently confuses left and right body parts. Inspired by \cite{yang2023set}, we incorporate a visual prompt to assist the model in distinguishing between left and right body parts. Specifically, we utilize ViTPose \cite{xu2022vitpose} for body keypoint detection, and then visually differentiate left and right body parts with distinct colors on the image and explicitly specify them in the text prompt provided to GPT4V, as shown in Figure \ref{fig:rpe_benchmark}. The details of the query input are represented in Table \ref{tab:person_description_query}. After generating these descriptions, we manually refine them and create 250 question-answer pairs in the following format:
``\texttt{{USER}}:\texttt{<IMAGE>} \texttt{\{descriptions\_person\}}, \texttt{\small can you give the SMPL pose of this person?}  \texttt{{ASSISTANT}}: \texttt{\small Sure, it is} \texttt{<POSE>}."  Here, \texttt{\{descriptions\_person\}} represents the person description from a specific aspect. 

\begin{table*}\centering
\begin{minipage}{1.99\columnwidth}\vspace{0mm}    \centering
\begin{tcolorbox} 
    \centering
    \small
     \hspace{-6mm}
\begin{tabular}{p{0.99\columnwidth}}
 \begin{minipage}{0.99\columnwidth}\vspace{0mm}
\textbf{(a)} You serve as an AI visual analyst for image examination. Your input will be an image containing humans. Your task is to provide descriptions of this individual. Your analysis should focus on four attributes: the individual's overall behavior, shape, outfits, and detailed pose.
For the overall behavior, if this person is doing specific activities like yoga or sports, provide a detailed name. For the outfits, specify the color of the clothes. For the detailed pose, describe as detail as possible, looking into the torso, left, right arms, hands, and legs. To help you distinguish the left arms/legs from the right arms/legs, we have drawn the left body joints with green color, while the right body joints with red color.  Don't mention the lines/marks/joints color in your answer!
Please output the attributes (behavior, shape, outfits, and pose) as keys in a JSON file format, each value should be one or two sentences.
\newline

\textbf{(b)} You serve as an AI assistant. Your input will be a description of a person from four attributes: overall behavior, shape, outfits, and detailed pose. Your task is to understand the provided descriptions and then use your reasoning ability to generate one comprehensive short description in a manner that requires an advancing logical reasoning ability to understand and distinguish the correct individual. Remember, the comprehensive description should be shorter than 30 words and do not need to cover all the details, and require a strong reasoning ability to understand.

\end{minipage}
    \end{tabular}
    
\end{tcolorbox}
\vspace{-2mm}
\caption{Example to query GPT4 for person description.  Prompt (a) is used to request GPT4V for detailed behavior, shape, outfits, and pose descriptions. Prompt (b) then instruct GPT4 to integrate and summarize these elements into a comprehensive description.}
    \label{tab:person_description_query}
\end{minipage}
\end{table*}

\begin{figure*}[!h]
  \centering  
  \includegraphics[width=0.99\textwidth]{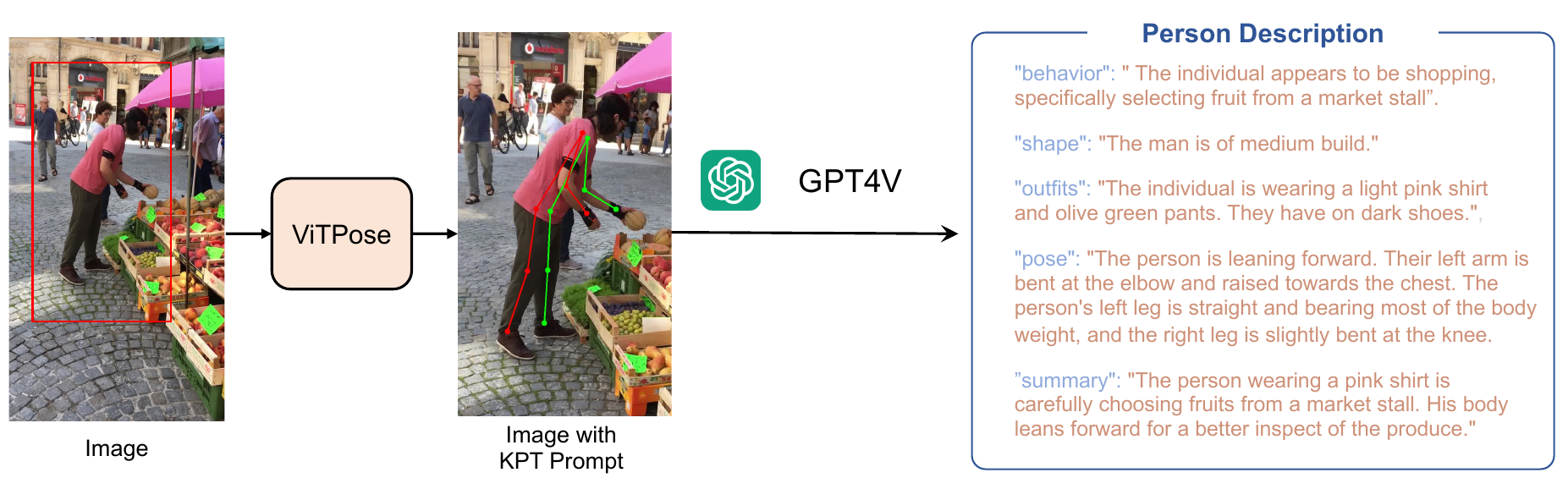}
  \caption{\small Illustration of our method to generate person descriptions for the RPE benchmark. We use ViTPose \cite{xu2022vitpose} to detect the body keypoints and mark the left-body and right-body joints with different colors as visual prompts, and then query GPT4V for descriptions.}
  \label{fig:rpe_benchmark}
\end{figure*}

\section{Ablation Study Details}
\paragraph{Representations of Human Pose.} Instead of utilizing the pose token \texttt{<POSE>}, an alternative approach to representing human poses involves using natural language, specifically textual descriptions specifying keypoint locations.  To facilitate a comparison between these two pose representations, we use the same dataset pairs as in \model and formulate Visual Question Answering (VQA) pairs for training. The question-answer template is structured as follows: ``\texttt{{USER}}: \texttt{<Image>} \texttt{\small There is a person in the image, please estimate the visible keypoints coordinates. The output format should be Nose:(x1,y1),Neck:(x2,y2),...}  \texttt{{ASSISTANT}: \small The detected visible keypoints are \{KEYPOINT\_NAME1\}:\{X1, Y1\}, \{KEYPOINT\_NAME2\}:\{X2, Y2\}, ...}". In this template, \texttt{<IMAGE>} represents the image patch token placeholder, \texttt{\small \{KEYPOINT\_NAME\}} denotes the name of the visible keypoint, and \texttt{\small \{X, Y\}} indicates the discretized keypoint coordinates. Figure \ref{fig:kpt_training_data} provides some examples of these training pairs. We then fine-tune the base model, LLaVA \cite{llava}, referred to as \llava*, to estimate keypoints and then use \smplify to transform the keypoints into a SMPL pose for comparison with our pose token \texttt{<POSE>} representation. Visual results of \llava* are displayed in Figure \ref{fig:llava_finetuned_result}. As shown, using textual descriptions as pose representation causes the network to often struggle to accurately estimate human poses and to often predict symmetrical poses, which may stem from the discretized nature of language signals.  

\begin{figure*}[!h]
  \centering  
  \includegraphics[width=0.96\textwidth]{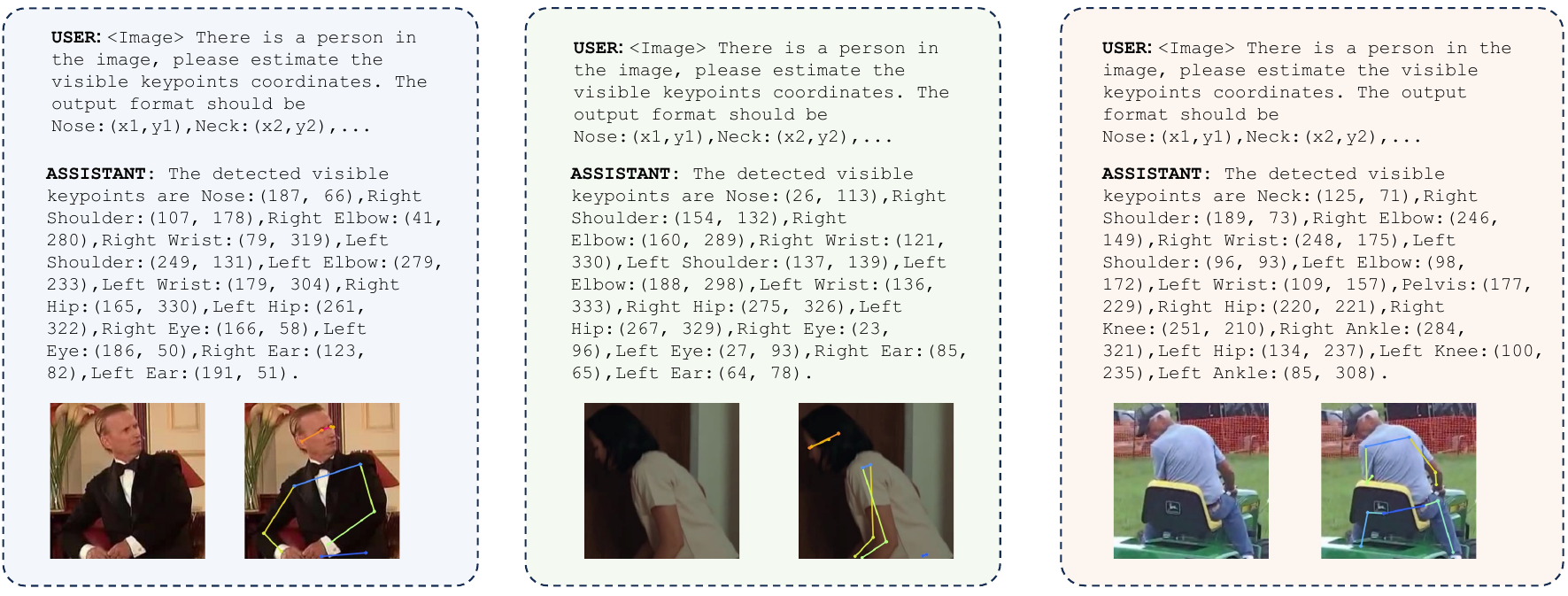}
  \vspace{-2mm}
  \caption{\small Examples of VQA data used to fine-tune the \llava model for pose estimation with textual descriptions of 2D keypoints.}
  \label{fig:kpt_training_data}
\end{figure*}

\begin{figure*}[!h]
  \centering  
  \includegraphics[width=0.96\textwidth]{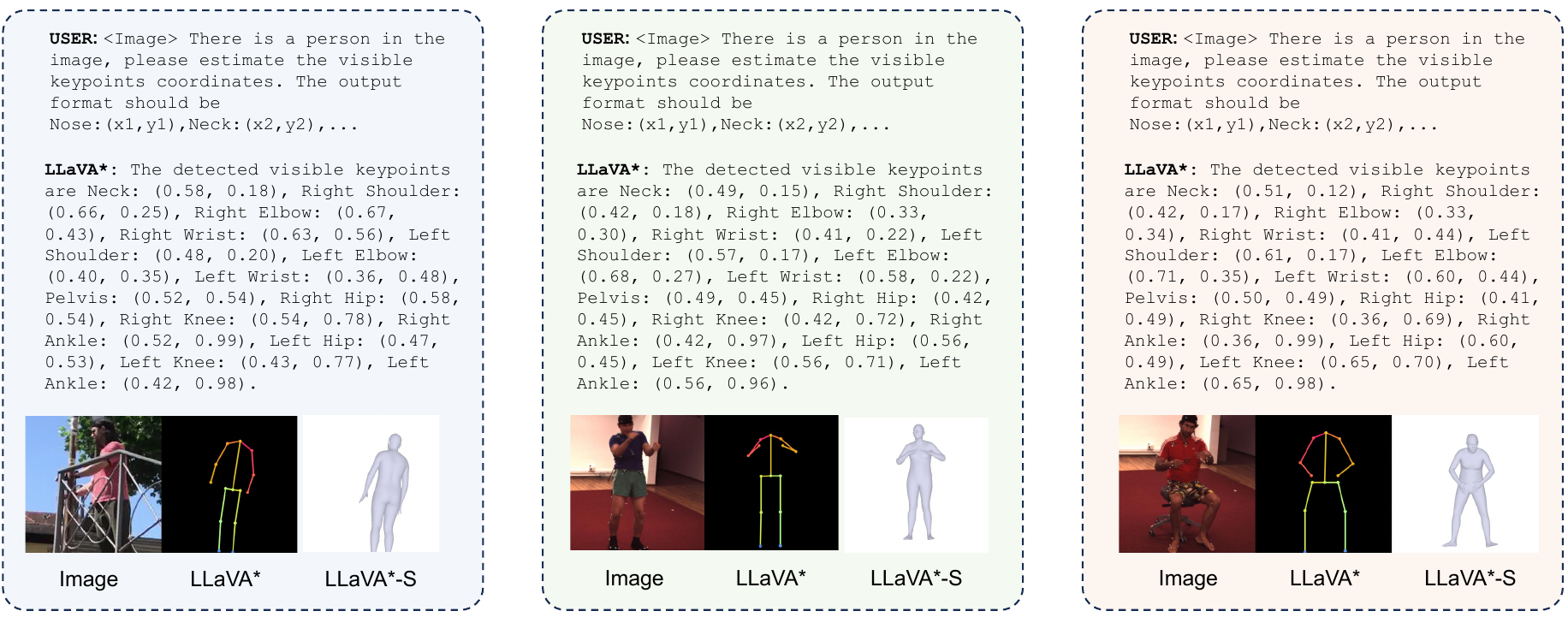}
  \vspace{-1mm}
  \caption{\small Visual results of \llava*. Given an RGB image, \llava* generates textual descriptions about keypoint locations. We then extract the keypoints from the textual descriptions and adopt \smplify~\cite{smplify} to fit the SMPL pose.}
  \label{fig:llava_finetuned_result}
  \vspace{-3mm}
\end{figure*}

\paragraph{Effects of Various Datasets.} 
\input{tables/08_datasets_ablation_v2}

\input{tables/09_backbone_ablation_v2}

For training, we utilize three data types: text-to-SMPL pose (Text2Pose), image-to-SMPL pose (Image2Pose), and general instruction-following data for visual question answer (VQA). To maintain the model's reasoning capabilities comparable to other LLMs, the VQA dataset is consistently used. For evaluating the effects of Text2Pose and Image2Pose, we fine-tune the model separately with each dataset.  Table \ref{tab:exp_dataset_ablation} presents the quantitative results.  
In contrast to the original LLaVA, which solely trains on VQA data, incorporating either Image2Pose or Text2Pose data into our model enhances pose estimation accuracy. Utilizing all data types, our model achieves optimal performance.

\paragraph{Multimodal LLM backbones.}
To evaluate how the LLM affects the performance of \model, we employ both the LLaVA-V1.5-7b\footnote{\href{https://huggingface.co/liuhaotian/llava-v1.5-7b}{\tt liuhaotian/llava-v1.5-7b}} and LLaVA-V1.5-13B\footnote{\href{https://huggingface.co/liuhaotian/llava-v1.5-13b}{\tt liuhaotian/llava-v1.5-13b}} models, which are based on the LLaMA-7b and LLaMA-13b backbones, respectively.  
Table~\ref{tab:exp_backbone_ablation} shows the comparisons between 7b and 13b models. 
The 13b model, despite needing more training time, delivers superior accuracy over the 7b model. This suggests that our method's effectiveness is contingent on the capabilities of the LLM models and also benefits from their rapid advancements.


\subsection{More Results}

\noindent\textbf{Generalization to Strong Occlusions.}
Even without any data augmentation during training, our model surprisingly still performs well on images with severe occlusions. Figure \ref{fig:hmr_severe_occlu} shows  pose estimation results for such cases. Even when half of the images are missing, \model can still produce reasonable human poses. 
This suggests that it is able to leverage its general visual knowledge about occlusion in solving the human pose estimation problem.

\begin{figure*}[!h]
  \centering  
  \includegraphics[width=0.85\textwidth]{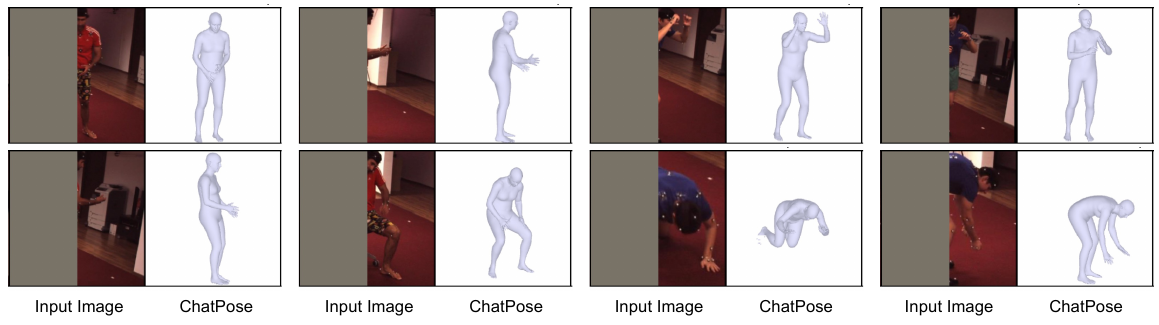}
  \vspace{-3mm}
  \caption{\small Pose estimation on images with significant occlusion. Without training for occlusion cases, \model is surprisingly robust.}
  \vspace{-3mm}
  \label{fig:hmr_severe_occlu}
\end{figure*}

\paragraph{Comparisons Details.} 
For pose estimation, when comparing with other multi-modal LLMs that do not directly output 3D human poses, we adopt two approaches: firstly, generating keypoint coordinates followed by SMPLify~\cite{smplify} optimization of the 3D pose, and secondly, producing textual descriptions of the pose that are then processed by PoseScript~\cite{posescript} to create SMPL pose parameters. The workflow for the first method is illustrated in Figure~\ref{fig:llava_finetuned_result}, and for the second method in Figure~\ref{fig:lmm_posescript}. 

\begin{figure*}[!h]
  \centering  
  \includegraphics[width=0.96\textwidth]{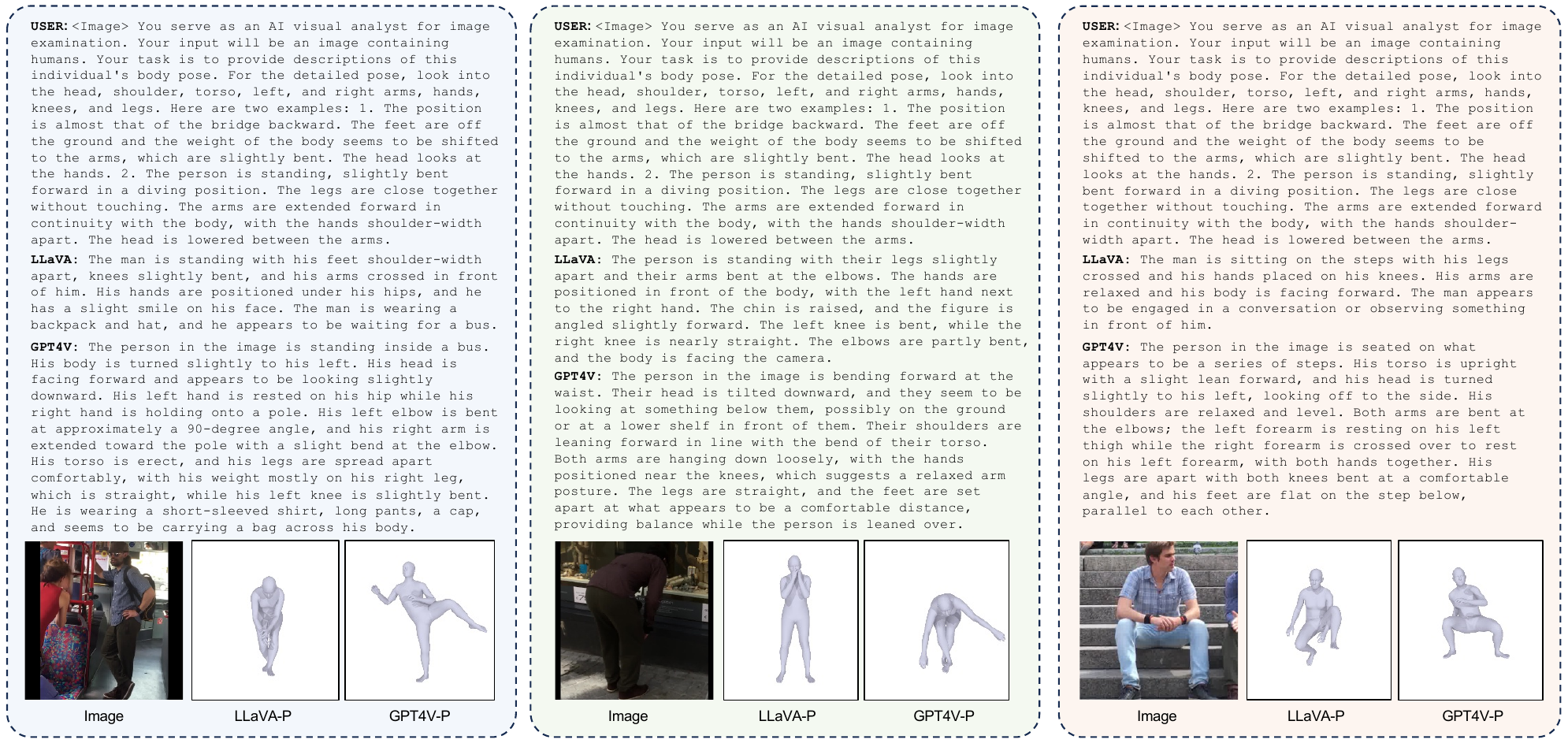}
  \vspace{-1.5mm}
  \caption{\small Visual results of \llava and GPT4. Given an RGB image, \llava and GPT4 generate textual descriptions about human poses. We then use PoseScript~\cite{posescript} to generate SMPL poses based on the text descriptions. }
  \label{fig:lmm_posescript}
\end{figure*}

\begin{figure*}[!h]
  \centering  
  \includegraphics[width=0.85\textwidth]{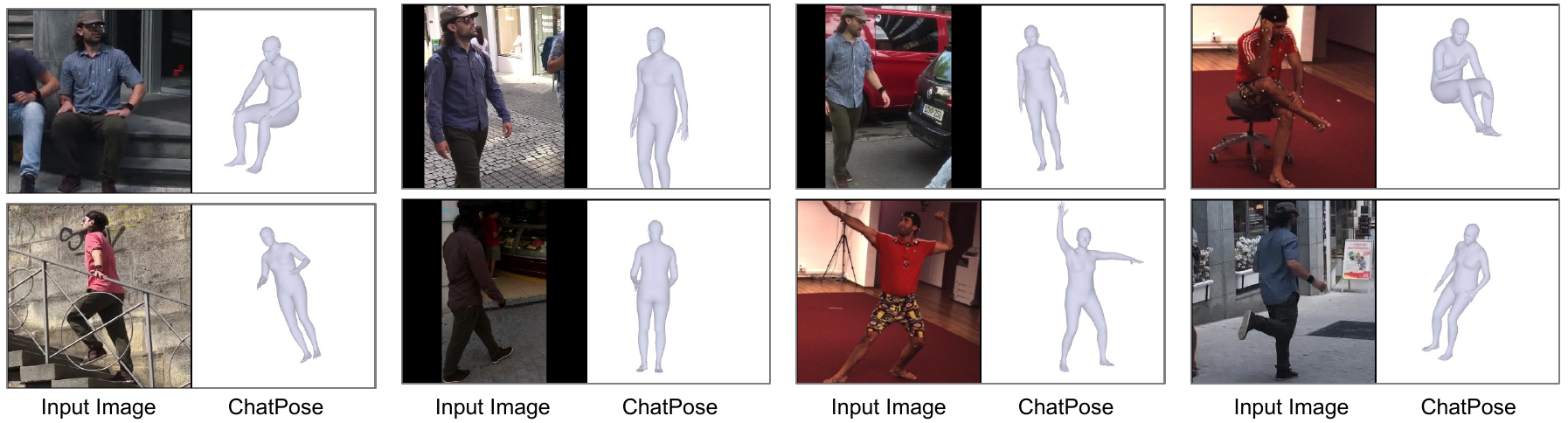}
  \vspace{-1.5mm}
  \caption{\small Failures cases of \model on the human pose estimation task. Note that a common failure mode is to estimate the articulated pose correctly but to output the incorrect global orientation.  }
  \vspace{-2mm}
  \label{fig:hmr_failure_cases}
\end{figure*}

\CR{
\noindent\textbf{FID for pose generation.} 
We evaluated FID on real poses from the PoseScript and 3DPW test sets, generating text descriptions for the latter using PoseScript Rules; see Tab.~\ref{tab:rebuttal_fid}.  FID reflects distribution similarity more than generation quality. Since PoseScript trains only on its data and our model uses data from PoseScript and HMR (w/o text); the scores reflect this. 
} 
\input{rebuttal/table_FID}

\CR{
\noindent\textbf{More analysis of T2P results} 
As shown Table 1 in main paper, \model lags behind for classical pose-to-text (P2T) retrieval while being on par with PoseScript~\cite{posescript} for classical text-to-pose (T2P) retrieval. We delve deeper into this analysis here. 
We start by visualizing instances where \model underperforms while PoseScript succeeds, with one such example illustrated in Figure~\ref{fig:rebuttal_fid}. Further analysis of failures did not reveal a distinct pattern. 
The contributing factors include: 1) Training strategy differences – PoseScript employs a VAE model with KL loss to ensure relative symmetry for T2P and P2T, whereas we employ LLMs with inherent strong priors about languages. 2) Varied training data – Unlike PoseScript's consistent use of AMASS, our multi-modal training employs a mix of AMASS, HMR, and general VQA data, leading to a varied training-test distribution. 
3) Bias in the retrieval models with P2T being less accurate than T2P (as noted in the PoseScript paper Tab.~1).  
We reevaluated P2T and T2P using a higher-accuracy retrieval model from the PoseScript journal version.  Top 5/10/50/100 P2T and T2P results are detailed in Tab.~\ref{tab:rebuttal_t2p}.  
}
\input{rebuttal/table_T2P}

\begin{figure}[!h]
  \centering  
  \includegraphics[width=0.38\textwidth]{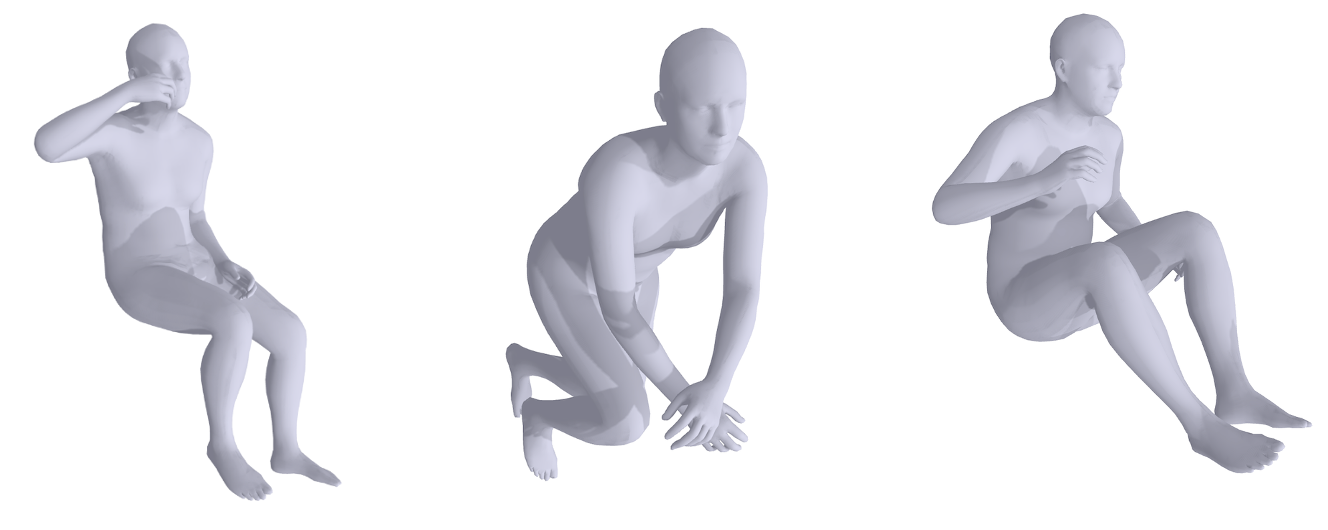}
  \vspace{-3mm}
  \caption{From left to right: GT, PoseScript, \model. This illustrates a comparison in pose generation between PoseScript and our approach. In instances where T2P retrieval is correct, PoseScript's P2T is also correct, whereas \model's P2T is incorrect. 
  }
  \vspace{-4mm}
  \label{fig:rebuttal_fid} 
\end{figure}

\CR{
\noindent\textbf{Other baselines for RPE and SPG.}
We show more baselines in Table \ref{tab:rebuttal_rpe}. 
Using LLaVA/GPT4 to convert SPG texts into PoseScript texts (LLaVA/GPT4+PoseScript) preforms poorly. 
To improve results we add in-context learning (w/ ICL) but this remains less accurate than \model.  
We finetuned PoseScript with SPG data; the results in are also less accurate than \model. 
}

\paragraph{Failure Cases.}
We also show some limitations of the current model in Figure \ref{fig:hmr_failure_cases}.
It is important to note that the global orientation can be significantly off, even when the body pose is approximately correct.
This global orientation issue might be improved by using a 
superior vision backbone, particularly one that excels at localization.

\input{rebuttal/table_RPE}
\newpage

%% file: tables/08_datasets_ablation_v2.tex
\begin{table}
	\centering
	\scriptsize
	\resizebox{0.48\textwidth}{!}{
		\begin{tabular}{|c|c|c|c|c|c|c|}
			\toprule
			\multirow{2}[4]{*}{\textbf{Method}} & \multirow{2}[4]{*}{VQA \cite{llava}} & \multirow{2}[4]{*}{Image2Pose} & \multirow{2}[4]{*}{Text2Pose} & \multicolumn{2}{c|}{ Pose Estimation} & \multirow{2}[4]{*}{\makecell{Reasoning-based\\Pose Estimation}} \\
			\cmidrule(lr){5-6}
			& & & & 3DPW \cite{threedpw} & H36M \cite{human36m} & \\
			\midrule
                LLaVA-P & \checkmark& & & 172.3 & 172.5 & 186.8 \\ 
                \model w/o Image2Pose & \checkmark& & \checkmark & 115.1& 121.6 & 123.7 \\			
               \model w/o Text2Pose & \checkmark & \checkmark & & 87.8& 89.2 & 109.8 \\
			\model full data & \checkmark& \checkmark & \checkmark& \textbf{81.9} & \textbf{82.4} & \textbf{101.8} \\
			\bottomrule
		\end{tabular} 
	}
	\caption{
	            \textbf{Ablation study: effect of different training data. }
				PA-MPJPE (in mm) is reported. Lower is better. 
	}
  \vspace{-4mm}
	\label{tab:exp_dataset_ablation}
\end{table}

%% file: tables/09_backbone_ablation_v2.tex
\begin{table}[t]
	\centering
	\scriptsize
	\resizebox{\columnwidth}{!}{
		\begin{tabular}{c|c|c|c|}
			\toprule
			\multirow{2}[4]{*}{Pretrained Model} & \multicolumn{2}{c|}{ Pose Estimation} & \multirow{2}[4]{*}{\makecell{Reasoning-based\\Pose Estimation}}  \\
			\cmidrule(lr){2-3}
			& 3DPW \cite{threedpw} & H36M \cite{human36m} &  \\
			\midrule
			LLaVA-V1.5-7B~\cite{liu2023improvedllava} & 84.5& 82.9 & 102.5 \\
			LLaVA-V1.5-13B~\cite{liu2023improvedllava} & 81.9 & 82.4 & 101.8  \\
			\bottomrule
		\end{tabular} 
	}
	\caption{
	            \cheading{Ablation study: effect of multimodal LLM backbones} 
             PA-MPJPE (in mm) is reported. Lower is better. 
	}
  \vspace{-4mm}
	\label{tab:exp_backbone_ablation}
\end{table}

%% file: rebuttal/table_FID.tex
\begin{table}[h]
	\centering
	\scriptsize
	\resizebox{0.85\columnwidth}{!}{
		\begin{tabular}{c|c|c}
			\toprule
			{Method} & FID (PoseScript) $\downarrow$ & FID (3DPW) $\downarrow$  \\ 
			\midrule
			PoseScript & \textbf{0.50}  & 1.21 \\
			\model  & 1.51 & \textbf{0.75}  
 \\  
			\bottomrule
		\end{tabular} 
	}
  \vspace{-1mm}
	\caption{
	    FID Scores on PoseScript and 3DPW dataset.  
	}
  \vspace{-4mm}
	\label{tab:rebuttal_fid}
\end{table}


%% file: rebuttal/table_T2P.tex
\begin{table}[h]
  \vspace{-2mm}
	\centering
	\scriptsize
	\resizebox{0.85\columnwidth}{!}{
		\begin{tabular}{c|c|c}
			\toprule
			{Method} & {$R^{P2T}$}  $\uparrow$ & {$R^{T2P}$} $\uparrow$ \\ 
			\midrule
			PoseScript & \textbf{22.6/31.0/57.9}/70.8 & 22.4/32.1/58.7/71.5\\
			\model  & 17.6/25.3/57.6/\textbf{71.2} & \textbf{28.0/39.0/70.4/83.5}
 \\  
			\bottomrule
		\end{tabular} 
	}
  \vspace{-1mm}
	\caption{
	    TOP 5/10/50/100 T2P and P2T results with retrieval model from PoseScript journal version. 
	}
  \vspace{-2mm}
	\label{tab:rebuttal_t2p}
\end{table}

%% file: rebuttal/table_RPE.tex
\begin{table}[t]
	\centering
	\scriptsize
	\resizebox{0.9\columnwidth}{!}{
		\begin{tabular}{|c|c|c|}
			\toprule
			Method & SPG {$R^{P2T}$} $\uparrow$  & SPG {$R^{T2P}$} $\uparrow$   \\
			\midrule
			LLaVA-P & 5.0/8.6/13.8 & 5.8/9.7/14.7 \\
			LLAVA-P (w/ ICL) & 2.6/5.3/9.2 & 3.5/6.3/10.5 \\
			GPT4-P & 3.5/6.9/11.3 & 4.1/7.3/11.9 \\
			GPT4-P (w/ ICL) & 3.7/7.6/13.1 & 5.1/8.1/13.5 \\
			PoseScript finetuned with SPG & 6.0/9.6/15.4 & 7.4/12.1/18.5 \\
			\model (ours) & 8.6/14.2/20.8 & 10.9/16.9/25.3 \\ 
			\bottomrule
		\end{tabular} 
	}
  \vspace{-1mm}
	\caption{
	    Results of suggested baselines. ICL means ``in context learning", where we teach LLaVA/GPT4  with a few examples of converting our SPG text to more detailed PoseScript descriptions. 
	}
  \vspace{-4.5mm}
	\label{tab:rebuttal_rpe}
\end{table}